\documentclass[letterpaper]{article} 
\usepackage{amsmath,amssymb}
\usepackage{bm}

\usepackage[preprint]{aaai2027}  
\usepackage[hyphens]{url}  
\usepackage{graphicx} 
\usepackage{natbib}  
\usepackage{caption} 
\usepackage{algorithm}
\usepackage{algorithmic}

\usepackage{newfloat}
\usepackage{listings}
\DeclareCaptionStyle{ruled}{labelfont=normalfont,labelsep=colon,strut=off} 
\floatstyle{ruled}
\newfloat{listing}{tb}{lst}{}
\floatname{listing}{Listing}

\usepackage{booktabs}

\usepackage{array, multirow,pifont}
\usepackage{threeparttable, makecell}
\usepackage{subcaption}
\newcommand{\cmark}{\ding{51}}
\newcommand{\xmark}{\ding{55}}
\newcolumntype{C}[1]{>{\centering\arraybackslash}p{#1}}

\newcommand{\MGD}{\scriptscriptstyle\mathrm{MGD3}}
\newcommand{\SSL}{\scriptscriptstyle\mathrm{SSL}}
\newcommand{\SRG}{\scriptscriptstyle\mathrm{SRG}}

\title{Self-Supervised Representation-Guided Generative Dataset Distillation}
\author{
Mingzhuo Li\textsuperscript{\rm 1},
Guang Li\textsuperscript{\rm 1}\thanks{Correspondence to Guang Li <guang@lmd.ist.hokudai.ac.jp>},
Linfeng Ye\textsuperscript{\rm 2},
Jiafeng Mao\textsuperscript{\rm 3},
Takahiro Ogawa\textsuperscript{\rm 1},\\
Konstantinos N. Plataniotis\textsuperscript{\rm 2},
and Miki Haseyama\textsuperscript{\rm 1}
}

\affiliations{
\textsuperscript{\rm 1}Hokkaido University, Japan\\
\textsuperscript{\rm 2}University of Toronto, Canada\\
\textsuperscript{\rm 3}University of Tokyo, Japan\\
\{mingzhuo, guang, ogawa, mhaseyama\}@lmd.ist.hokudai.ac.jp\\
linfeng.ye@mail.utoronto.ca,
kostas@ece.utoronto.ca,
mao@hal.t.u-tokyo.ac.jp
}

\begin{document}

\maketitle

\begin{abstract}
Dataset distillation compresses a large training set into a compact synthetic set while retaining its downstream utility. Most existing methods target randomly initialized networks, whereas modern vision systems often adapt frozen pretrained encoders with lightweight modules. Distilled samples should therefore preserve the discriminative geometry of the pretrained representation space, which existing generative objectives do not explicitly consider. We propose self-supervised representation-guided generative dataset distillation (SRG), a framework that translates the SSL geometry into diffusion guidance. Specifically, SRG constructs class-wise prototypes from real-image SSL representations and performs guidance through three SSL-space objectives for prototype alignment, inter-class discrimination, and intra-class assignment. During diffusion sampling, it adopts a stage-wise guidance strategy: early denoising is anchored to the latent of the real image whose SSL representation is nearest to the assigned prototype, whereas later denoising is guided by the SSL-space objectives. This division preserves the visual realism provided by the generative prior while progressively steering samples toward representative and class-discriminative regions of the SSL representation space. SRG consistently outperforms the evaluated generative baselines across multiple datasets and IPC settings. A cross-encoder evaluation further indicates transfer across pretrained representation spaces. These results demonstrate the effectiveness of representation-guided generation for dataset distillation with pretrained SSL models. 
\end{abstract}


\section{Introduction}

The rapid progress of computer vision has been supported by increasingly large models and training datasets \citep{schmidhuber2015NNdlReview,talaei2023review}. However, repeatedly storing, processing, and accessing such datasets incurs substantial computational and storage costs, particularly when models must be trained or evaluated across multiple experimental settings \cite{strubell2019cost}. Dataset distillation (DD) \citep{wang2018datasetdistillation,li2022awesome} provides a potential solution by compressing a large dataset into a much smaller synthetic set. Ideally, models trained on the distilled dataset should retain much of the performance achieved using the original data, thereby reducing the cost of repeated model training.

\par

\begin{figure}[t]
    \centering
    \includegraphics[width=\linewidth]{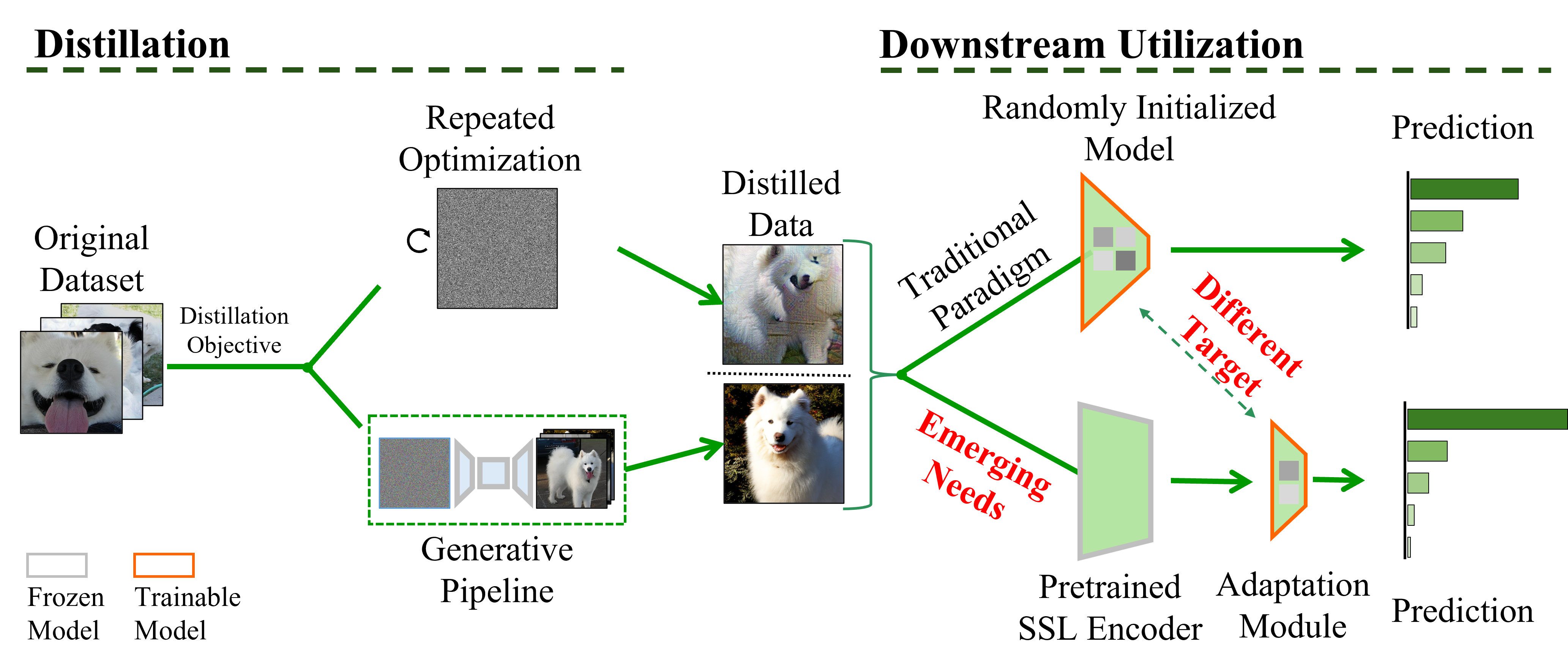}
    \caption{Overview of dataset distillation and downstream utilization. Traditional DD targets training from random initialization, whereas adaptation with pretrained SSL encoders requires distilled samples that preserve discriminative representation structure.}
    \label{fig:intro}
\end{figure}

As illustrated on the left side of Fig.~\ref{fig:intro}, existing DD methods can be broadly divided into non-generative and generative approaches. Non-generative methods directly optimize synthetic data as free variables, typically by matching training-related information, such as model gradients \citep{zhao2021DC, zhao2021DSA, li2023ddpp, li2024iadd}, between the real and distilled datasets. In contrast, generative DD methods construct distilled samples by incorporating distillation objectives into a pretrained generative pipeline \cite{li2024generative, su2024diffusion, wu2025dc3, ye2025igds, zou2025vlcp, cai2026evlf}. By generating samples on demand and avoiding the direct optimization of a fixed set of synthetic data, generative methods provide a flexible way to construct distilled datasets for different images-per-class (IPC) budgets.

\par

As shown on the right side of Fig.~\ref{fig:intro}, DD has traditionally been formulated for training randomly initialized models. Modern vision systems, however, increasingly reuse pretrained self-supervised learning (SSL) models as fixed feature encoders and train only lightweight task-specific modules, such as linear probes, on the extracted representations \citep{chen2020sslDownstream,gui2024sslReviewNew}. This shift changes the objective of dataset distillation. Rather than reproducing the complete training dynamics of a model trained from scratch, distilled samples should provide compact supervision that supports effective decision boundaries in a fixed representation space. Therefore, they need to preserve not only visual characteristics but also the representative and discriminative structure emphasized by the pretrained SSL encoder.

\par

Most existing DD methods are not explicitly designed for this objective. LGM \citep{cazenavette2025lgm} optimizes synthetic samples by matching gradients induced by a downstream linear classifier, while CLP-DD \citep{peng2026clpdd} improves efficiency using a closed-form solver and a discriminative outer objective. Although these methods demonstrate the feasibility of DD for pretrained encoders, both optimize a fixed synthetic set for each dataset and IPC budget, requiring a new optimization process when either changes. Generative DD offers a more flexible alternative, but existing objectives are mainly designed for training models from scratch. For example, MGD3 \citep{chan-santiago2025mgd3} guides diffusion sampling toward modes obtained in the generator's latent space. These modes capture visual variation learned by the generator but do not explicitly retain the discriminative geometry of the downstream SSL encoder.

\par

To address this gap, we propose self-supervised representation-guided generative dataset distillation (SRG), a framework that translates the SSL geometry into distillation signals to guide diffusion sampling. SRG constructs class-wise prototypes from real-image representations and assigns each denoising trajectory to one prototype. It then applies stage-wise guidance to incorporate both visual and representation-level information. During early denoising stages, latent-space guidance anchors the trajectory to a real-image latent associated with the assigned prototype. During later denoising stages, SSL-space guidance refines the generated representation through prototype alignment, inter-class discrimination, and intra-class assignment. These objectives align generated samples with representative regions, separate them from other classes, and encourage distinct assignments within each class. In this way, SRG directly incorporates downstream representation structure into diffusion sampling, obtaining distilled datasets tailored for SSL models.

\par

Our main contributions are summarized as follows:
\begin{itemize}
    \item We propose SRG, a framework that incorporates the discriminative geometry of pretrained SSL representations into diffusion sampling for dataset distillation.
    
    \item We develop a stage-wise guidance strategy that combines early latent-space anchoring with later SSL-space guidance using prototype alignment, inter-class discrimination, and intra-class assignment.
    
    \item We conduct extensive experiments across multiple datasets, IPC settings, and SSL encoders, demonstrating the effectiveness, scalability, and cross-encoder transfer of SRG.
\end{itemize}

\section{Related Work}

\subsection{Pretrained Vision Encoders}

Self-supervised learning (SSL) learns transferable visual representations from large-scale data without requiring manually annotated class labels \cite{gui2024sslReviewNew,jaiswal2021sslReviewLearn}. Pretrained encoders such as CLIP \cite{radford2021clip}, DINO \cite{caron2021dino}, MoCo \cite{he2020moco}, and EVA \cite{fang2023eva} have become widely used backbones for visual recognition. A common adaptation strategy is to freeze the encoder and train only a lightweight task-specific module on the extracted representations \cite{chen2020sslDownstream}. Under this setting, the utility of a compact training set depends on whether its samples preserve representative and class-discriminative regions in the pretrained representation space, motivating representation-aware dataset distillation.

\subsection{Dataset Distillation}
Dataset distillation (DD) \cite{wang2018datasetdistillation} 
synthesizes a compact dataset that retains the training utility of a much larger original dataset. DD has been applied to various downstream tasks, including fine-grained recognition \cite{ma2026fd2}, multimodal learning \cite{li2025davdd}, and privacy-preserving learning \cite{li2020soft, li2022compressed, li2023sharing}. However, most existing methods are designed for training randomly initialized models \cite{yu2023DDreview, liu2025DDreview}, although recent studies have begun to consider downstream adaptation with frozen pretrained encoders.

\paragraph{Non-generative methods.}
Non-generative DD methods typically optimize a fixed set of synthetic images by matching model gradients \cite{zhao2021DC}, training trajectories \cite{cazenavette2022trajMatch}, feature statistics or distributions \cite{wang2022featMatch, zhao2023distMatch, li2025hdd}, or kernel-based quantities \cite{nguyen2021kernal}. LGM \cite{cazenavette2025lgm} extends gradient matching to pretrained encoders using gradients from a downstream linear classifier together with Differentiable Siamese Augmentation \cite{zhao2021DSA}. CLP-DD \cite{peng2026clpdd} instead employs a closed-form linear-probe solver with a discriminative outer objective. Although both methods achieve strong performance, they optimize a fixed synthetic set for each target dataset and IPC budget, generally requiring re-optimization when either setting changes.

\paragraph{Generative methods.}
Generative DD methods exploit pretrained generative models to reduce reliance on directly optimizing a fixed synthetic set. IT-GAN \cite{zhao2022ganDD} optimizes informative latent codes of a frozen generative adversarial network, while Minimax \cite{gu2024minimax} incorporates representativeness and diversity objectives into the fine-tuning of generative models. MGD3 \cite{chan-santiago2025mgd3} and IGD \cite{chen2025igd} modify diffusion sampling using latent-mode and trajectory-influence guidance, respectively, whereas CaO2 \cite{wang2025cao2} refines generated samples through post-processing. However, these methods primarily target models trained from scratch and do not explicitly incorporate the representation geometry of a pretrained SSL encoder. SRG addresses this gap by incorporating SSL-space prototypes and objectives as distillation signals to guide diffusion sampling.

\begin{figure*}[t]
    \centering
    \includegraphics[width=0.9\linewidth]{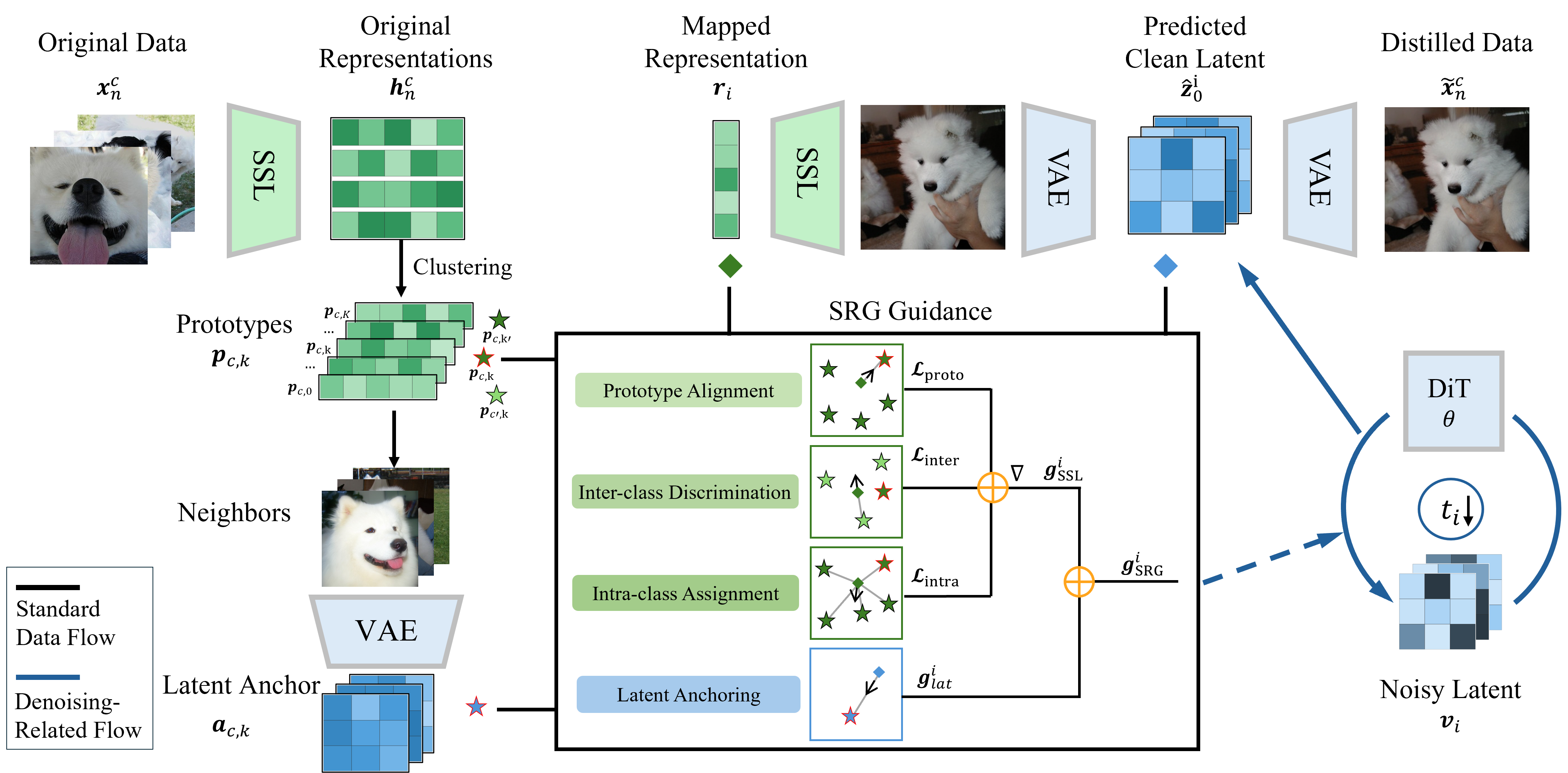}
    \caption{
    Overview of SRG. Class-wise prototypes are constructed by clustering real-image representations extracted by a frozen SSL encoder, and each denoising trajectory is assigned to one target prototype. Latent-space guidance anchors the trajectory to a prototype-associated real-image latent, while SSL-space guidance improves prototype alignment, inter-class discrimination, and intra-class assignment.}
    \label{fig:workflow}
\end{figure*}

\section{Method}

\subsection{Preliminaries}
We employ a pretrained diffusion transformer (DiT) \cite{Peebbles2023DiT} that operates in the latent space of a variational autoencoder (VAE) \cite{kingma2013VAE}. Starting from random noise, DiT generates images through iterative denoising.

\par

Let $\bm{v}_i$ denote the noisy latent at the $i$-th denoising step, corresponding to diffusion timestep $t_i$. Given the class condition $\bm{c}$, the pretrained DiT parameterized by $\theta$ predicts the noise component in $\bm{v}_i$.
The next latent $\bm{v}_{i+1}$ is obtained following the Gaussian reverse transition:
\begin{equation}
\bm{v}_{i+1}
=
\mu_{\theta}(\bm{v}_i,t_i,\bm{c})
+
\sigma_{t_i} \bm{\epsilon}_i, 
\label{eq:dit_update}
\end{equation}
where $\mu_{\theta}(\bm{v}_i,t_i,\bm{c})$ is the conditional mean of the next latent computed by the DiT sampler. $\sigma_{t_i}$ is the standard deviation of the reverse transition, and $\bm{\epsilon}_i$ is the Gaussian noise sampled at the $i$-th step. At each step, the sampler also estimates the clean latent $\hat{\bm{z}}_0^i$, which is decoded by VAE to the distilled image $\tilde{\bm{x}}_n^c$ after denoising.

\par
Although DiT can generate realistic images, the standard sampling process does not explicitly incorporate a distillation objective. MGD3 \cite{chan-santiago2025mgd3} addresses this limitation by discovering representative latent modes in the original dataset and aligning each generated sample with a specific mode. Given a target budget of $\mathrm{IPC}=K$, MGD3 performs $K$-means clustering in the VAE latent space separately for each class $c$ to obtain a set of latent modes $\mathcal{M}_c =  \{\bm{m}_{c,k}\}_{k=1}^{K} $. Each generated sample is assigned to one mode $\bm{m}_{c,k}$ so that the generated set covers multiple regions of the original data distribution. For clarity, we omit the indices $(c,k)$ from the trajectory variables in the following formulations, and define the MGD3 guidance term as follows:
\begin{equation}
\label{eq:g_mgd3}
    \bm{g}_{\MGD}^{i}
    = - \lambda_{\mathrm{lat}}\sigma_{t_i}
    \left( \hat{\bm{z}}_0^{\,i}
    - \bm{m}_{c,k} \right), 
\end{equation}
where $\lambda_{\mathrm{lat}}$ controls the guidance strength, and $\sigma_{t_i}$ is used for timestep-dependent scaling.
The guidance is added to the standard reverse transition in Eq.~\ref{eq:dit_update}:
\begin{equation}
    \bm{v}_{i+1}
    = \mu_{\theta}(\bm{v}_i,t_i,\bm{c})
    + \sigma_{t_i}\bm{\epsilon}_i
    + \bm{g}_{\MGD}^{i}.
\end{equation}
By assigning different generation trajectories to distinct latent modes, MGD3 is designed to improve the coverage of the distilled dataset. However, using modes constructed in the latent space, its guidance does not explicitly preserve the discriminative structure required for downstream adaptation of pretrained SSL encoders.

\subsection{Prototype Alignment}
As shown in Fig.~\ref{fig:workflow}, we address this limitation by complementing latent-space anchoring with guidance objectives derived from the SSL representation space. Specifically, we construct class-wise SSL prototypes and enable guidance through differentiable objectives whose gradients are backpropagated to the predicted clean latent. Unlike latent modes used in MGD3, which characterize the visual distribution encoded by the generator, SSL prototypes summarize the structure emphasized by the SSL encoder, providing guidance that better aligns with the downstream adaptation setting. For each real image $\bm{x}_n^c$ from class $c$, we extract the $\ell_2$-normalized representation $\bm{h}_n^c=\operatorname{norm}_2(f_{\phi}(\bm{x}_n^c))$ using the frozen SSL encoder $f_{\phi}$, where $\operatorname{norm}_2(\bm{a})=\bm{a} / \|\bm{a}\|_2$. Normalization removes feature-magnitude effects, avoiding encoder-specific magnitude calibration in the guidance objectives. The class-wise prototypes are obtained by applying spherical $K$-means within each class as follows:
\begin{equation}
\left \{ \bm{p}_{c,k} \right\}_{k=1}^{K}
= \operatorname{SphKMeans}
\left (\left\{\bm{h}_n^c\right\}_{n=1}^{N_c}; K
\right ), 
\end{equation}
where $\bm{p}_{c,k}$ denotes the normalized $k$-th prototype of class $c$ and $N_c$ denotes the number of images in class $c$. 
Each prototype represents a local region of the class distribution in the SSL representation space. Accordingly, each generation trajectory is assigned to one prototype.

\par

However, the predicted clean latent $\hat{\bm{z}}_0^{i}$ and the SSL prototype $\bm{p}_{c,k}$ lie in different spaces and cannot be compared directly. To solve this problem, we map the predicted clean latent to an SSL representation $\bm{r}_i = \operatorname{norm}_2(f_{\phi}(D_{\psi}(\hat{\bm{z}}_0^{i})))$, where $D_{\psi}$ denotes the VAE decoder. Since both $\bm{r}_i$ and $\bm{p}_{c,k}$ are normalized, their cosine similarity can be measured by:
\begin{equation}
\operatorname{sim}(\bm{a}, \bm{b}) = \bm{a}^T \bm{b}.
\end{equation}
To convert this SSL-space similarity into guidance for diffusion sampling, we formulate differentiable representation objectives and use their gradient as guidance directions. First, we introduce the following prototype alignment objective to drive the mapped representation of the generated sample toward the assigned prototype:
\begin{equation}
\mathcal{L}_{\mathrm{proto}}^i
= 1 - \operatorname{sim} \left(\bm{r}_i, \bm{p}_{c,k} \right ).
\end{equation}
Minimizing this objective aligns each generated representation with a local, representative region of the corresponding class in the SSL representation space.

\subsection{Inter-Class Discrimination}
The prototype alignment objective pulls each generated representation toward its assigned prototype. However, for semantically similar classes, an assigned prototype may lie near regions occupied by neighboring classes, allowing the generated representation to cross the class boundary. To explicitly preserve separation, we introduce an inter-class discrimination objective. Specifically, we measure the similarity between the generated representation $\bm{r}_i$ and all prototypes assigned to other classes. Their effects are aggregated using log-sum-exp, which places greater emphasis on highly similar prototypes while retaining contributions from the remaining ones. The objective is defined as follows:
\begin{equation}
\mathcal{L}_{\mathrm{inter}}^i = 
\log \sum_{\substack{c'=1 \\ c'\neq c}}^{C} \sum_{k'=1}^{K}
\exp
\left(
\frac{\operatorname{sim}
\left(
\bm{r}_{i}, \ \bm{p}_{c',k'}
\right)
}{\tau_{\mathrm{inter}}}
\right), 
\end{equation}
where $\tau_{\mathrm{inter}}$ is a temperature parameter controlling the concentration of the objective. A smaller temperature makes the objective more sensitive to the most similar prototype, whereas a larger temperature distributes weight more evenly across prototypes. 
Minimizing $\mathcal{L}_{\mathrm{inter}}^i$ reduces similarity between the generated representation and prototypes of other classes, thereby encouraging class-level separation in the SSL representation space.

\subsection{Intra-Class Assignment}
Despite these efforts, multiple generation trajectories of the same class may still converge to similar regions, particularly when the pretrained generator strongly favors certain visual patterns. This tendency can weaken correspondence with the assigned prototypes in higher-IPC settings. We therefore introduce an intra-class assignment objective to make the generated representation more similar to the assigned prototype than to other prototypes of the same class. For the generation trajectory assigned to $\bm{p}_{c, k}$, the objective is defined using a cross-entropy loss as follows:
\begin{equation}
    \mathcal{L}_{\mathrm{intra}}^i
    = - \log 
    \frac{
    \exp \left(
    \operatorname{sim} \left(
    \bm{r}_i,\bm{p}_{c, k} \right)
    / \tau_{\mathrm{intra}} \right) }
    { \displaystyle \sum_{k'=1}^{K}
    \exp \left(
    \operatorname{sim} \left(
    \bm{r}_i,\bm{p}_{c, k'} \right)
    / \tau_{\mathrm{intra}} \right) } ,
\label{eq}
\end{equation}
where $\tau_{\mathrm{intra}}$ is the temperature parameter controlling the sharpness of the prototype assignment. Minimizing $\mathcal{L}_{\mathrm{intra}}^i$ increases similarity to the assigned prototype relative to competing prototypes, thereby strengthening the assignment and discouraging different trajectories from converging to the same region.

\subsection{Overall Generation}
We combine the three representation objectives into the overall SSL-space guidance objective:
\begin{equation}
    \mathcal{L}_{\mathrm{SSL}}^{i} = \mathcal{L}_{\mathrm{proto}}^{i} + \mathcal{L}_{\mathrm{inter}}^{i} + \mathcal{L}_{\mathrm{intra}}^{i}.
\end{equation}
Since all representations are $\ell_2$-normalized, the similarity terms are defined on a consistent range. For simplicity, we use the same weights for all three objectives and rely on the temperature parameters to modulate the effects of the discrimination and assignment terms. The SSL-space guidance is obtained by backpropagating the objective to the predicted clean latent as follows:
\begin{equation}
    \bm{g}_{\SSL}^{i}
    = -\lambda_{\mathrm{SSL}}
    \nabla_{\hat{\bm{z}}_{0}^{i}}
    \mathcal{L}_{\mathrm{SSL}}^i, 
\end{equation}
where $\lambda_{\mathrm{SSL}}$ controls the guidance strength. 

\par

The reliability of the two guidance signals varies over the denoising stages. Early stages form image structure, where latent-space anchoring is effective yet the predicted latent is uncertain for stable semantic comparison. As denoising progresses, representation becomes increasingly discriminative \cite{kim2025diffTheory}, making SSL more informative and structural anchoring less critical. We therefore separate the two signals with the following schedule:
\begin{align}
\begin{split}
    \bm{g}_{\SRG}^{i} 
    = \mathbb{I}[t_i \geq t_{\mathrm{lat}}]
    \ \bm{g}_{\mathrm{lat}}^{i}
    \\
    + \mathbb{I}[t_i < t_{\mathrm{SSL}}] 
    \ \bm{g}_{\SSL}^{i}, 
\end{split}
\end{align}
where $\mathbb{I}[\cdot]$ denotes the indicator function, and $t_{\mathrm{lat}}$ and $t_{\SSL}$ determine the timesteps at which latent-space guidance stops and SSL-space guidance starts, respectively. 
The latent-space guidance $g_{\mathrm{lat}}^i$ follows Eq.~\ref{eq:g_mgd3} with $\bm{m}_{c,k}$ replaced by $\bm{a}_{c,k}$, the latent of the real image whose SSL representation is closest to $\bm{p}_{c,k}$ in cosine similarity, serving as a trajectory anchor rather than the guidance target in the original definition.

\par

\begin{table*}[t]
\centering
\caption{
Comparison of downstream validation accuracy between SRG and baseline methods on different ImageNet benchmarks using DINOv2. Random denotes random sampling from the original dataset, while Neighbor denotes selection of the real sample nearest to the assigned SSL prototype. The best result in each setting is highlighted in bold.
}
\label{tab:main}
\begin{threeparttable}
\begin{tabular}{lccccccccc}

\toprule
\multirow{2}{*}{Method} 
& \multicolumn{3}{c}{ImageNet-IDC} 
& \multicolumn{3}{c}{ImageNet-100} 
& \multicolumn{3}{c}{ImageNet-1K} 
\\

\cmidrule(lr){2-4} \cmidrule(lr){5-7} \cmidrule(lr){8-10}
& IPC=1 & IPC=3 & IPC=5 
& IPC=1 & IPC=3 & IPC=5 
& IPC=1 & IPC=3 & IPC=5 
\\

\midrule
Random 
& $85.6_{\pm 2.5}$ 
& $96.2_{\pm 0.8}$ 
& $97.0_{\pm 0.5}$

& $77.8_{\pm 0.3}$ 
& $86.2_{\pm 0.1}$ 
& $89.3_{\pm 0.1}$ 

& $53.7_{\pm 0.1}$ 
& $67.5_{\pm 0.1}$ 
& $70.8_{\pm 0.1}$ 
\\

Neighbor 
& $93.7_{\pm 0.7}$ 
& $96.9_{\pm 0.6}$ 
& $97.8_{\pm 0.4}$

& $86.8_{\pm 0.2}$ 
& $91.2_{\pm 0.1}$ 
& $92.5_{\pm 0.1}$ 

& $71.5_{\pm 0.1}$ 
& $75.6_{\pm 0.0}$ 
& $76.5_{\pm 0.1}$  
\\

DiT 
& $92.2_{\pm 0.9}$ 
& $95.2_{\pm 1.1}$ 
& $95.2_{\pm 0.7}$ 

& $82.8_{\pm 0.1}$ 
& $89.1_{\pm 0.4}$ 
& $89.0_{\pm 0.1}$ 

& $65.5_{\pm 0.2}$ 
& $71.2_{\pm 0.1}$ 
& $72.4_{\pm 0.0}$ 
\\

MGD3 
& $88.6_{\pm 1.2}$ 
& $95.6_{\pm 0.6}$ 
& $96.6_{\pm 0.5}$

& $81.7_{\pm 0.3}$ 
& $87.8_{\pm 0.1}$ 
& $90.2_{\pm 0.1}$ 

& $65.1_{\pm 0.0}$ 
& $72.0_{\pm 0.0}$ 
& $73.8_{\pm 0.1}$ 
\\

IGD
& $86.6_{\pm 2.5}$ 
& $94.2_{\pm 1.0}$ 
& $95.7_{\pm 0.8}$ 

& $82.7_{\pm 0.3}$ 
& $87.7_{\pm 0.2}$ 
& $89.6_{\pm 0.2}$

& $66.0_{\pm 0.1}$
& $71.9_{\pm 0.1}$
& $73.9_{\pm 0.0}$
\\

LGM
& $97.0_{\pm 0.5}$
& $98.2_{\pm 0.3}$ 
& $98.3_{\pm 0.2}$

& \bm{$91.0_{\pm 0.1}$} 
& $91.7_{\pm 0.1}$ 
& \textsc{OOM} 

& \color{gray}{$75.0_{\pm 0.1}$}\tnote{*}  
& \textsc{OOM} 
& \textsc{OOM} 
\\

CLP-DD
& \bm{$98.2_{\pm 0.2}$} 
& \bm{$98.4_{\pm 0.2}$}
& $98.4_{\pm 0.1}$

& $89.2_{\pm 0.1}$ 
& $91.3_{\pm 0.1}$ 
& $91.8_{\pm 0.0}$ 

& \color{gray}{$75.6_{\pm 0.1}$}\tnote{*}  
& \textsc{OOM} 
& \textsc{OOM} 
\\

SRG 
& $95.3_{\pm 0.4}$ 
& $98.1_{\pm 0.5}$ 
& \bm{$98.5_{\pm 0.4}$}

& $89.0_{\pm 0.1}$ 
& \bm{$92.3_{\pm 0.1}$} 
& \bm{$93.2_{\pm 0.2}$} 

& \bm{$73.5_{\pm 0.1}$}
& \bm{$76.3_{\pm 0.0}$} 
& \bm{$77.6_{\pm 0.1}$} 
\\

\midrule
Full 
& $99.7_{\pm 0.1}$ 
& $99.7_{\pm 0.1}$
& $99.7_{\pm 0.1}$

& $95.2_{\pm 0.1}$
& $95.2_{\pm 0.1}$
& $95.2_{\pm 0.1}$

& $83.0_{\pm 0.0}$
& $83.0_{\pm 0.0}$
& $83.0_{\pm 0.0}$

\\
\bottomrule
\end{tabular}

\begin{tablenotes}
\small
\item[*] We report results from the corresponding original papers because reproduction exceeded the available GPU memory.
\end{tablenotes}

\end{threeparttable}
\end{table*}

The final guided reverse transition is defined as follows:
\begin{equation}
    \bm{v}_{i+1}
    = \mu_{\theta}(\bm{v}_i,t_i,\bm{c})
    + \sigma_{t_i}\bm{\epsilon}_i
    + \bm{g}_{\SRG}^{i}.
\end{equation}
This formulation combines latent-space anchoring with SSL-space prototype alignment, inter-class discrimination, and intra-class assignment, yielding a DD framework tailored to pretrained SSL models. 
The stage-wise schedule uses the generator-facing signal when the trajectory is structurally uncertain and the encoder-facing signal after semantic content emerges. SRG thereby retains the visual plausibility of diffusion generation while explicitly steering samples toward the discriminative structure used for downstream adaptation. 

\section{Experiments}
\subsection{Experimental Settings}
For evaluation, we freeze each pretrained SSL encoder and repeat linear-probe training five times, with $1,000$ epochs per run. All evaluated datasets are derived from ImageNet-1K \cite{deng2009imageNet, deng2015imageNet1k}, a large-scale benchmark commonly used in dataset distillation. MGD3 \cite{chan-santiago2025mgd3}, IGD, and SRG use the same pretrained DiT model \cite{Peebbles2023DiT}. Unless otherwise specified, SRG uses $t_{\mathrm{lat}}=20$, $t_{\SSL}=20$,  $\lambda_{\mathrm{lat}}=0.1$, $\lambda_{\SSL}=15$, $\tau_{\mathrm{inter}}=1.0$, and $\tau_{\mathrm{intra}}=5.0$ across all datasets and IPC settings. Each SSL encoder uses a ViT-B \cite{dosovitskiy2021vit} backbone with the patch size
specified by the corresponding pretrained model. The prototypes are constructed on the GPU using an encoding batch size of $64$. All experiments are conducted on a single NVIDIA RTX A6000 GPU, except for the ImageNet-100 LGM experiment at $\mathrm{IPC} = 3$, which uses four A6000 GPUs due to OOM issues. Further implementation details are provided in the supplementary material.

\subsection{Benchmark Results}
We first compare SRG with representative methods on ImageNet-IDC \cite{kim2022IDC}, ImageNet-100 \cite{kim2022IDC}, and ImageNet-1K \cite{deng2015imageNet1k} under IPC settings of $1$, $3$, and $5$, using DINOv2 \cite{oquab2024dinov2} as the distillation and downstream-evaluation encoder. The compared methods include two selection-based baselines: random sampling from the original dataset (Random) and selection of the real sample nearest to each SSL prototype (Neighbor). For generative approaches, we include direct sampling from DiT \cite{Peebbles2023DiT}, the latent-guidance-based MGD3 \cite{chan-santiago2025mgd3}, and the influence-guidance-based IGD \cite{chen2025igd}. We further compare with SSL-oriented non-generative methods, including the pioneering LGM \cite{cazenavette2025lgm} and the computationally efficient CLP-DD \cite{peng2026clpdd}. We also report performance obtained with the full training set as an upper reference.

\par

As shown in Table~\ref{tab:main}, SRG achieves the best performance among the evaluated generative methods in every reported setting, supporting the effectiveness of incorporating SSL prototypes into generation for downstream adaptation of pretrained SSL encoders. Although optimization-based methods generally outperform at $\mathrm{IPC} = 1$, SRG closes the gap and attains better performance as IPC increases. We hypothesize that direct synthetic-image optimization gives LGM and CLP-DD greater freedom to encode various data patterns into each image, which is especially advantageous under a very small IPC budget. By contrast, SRG constrains samples to the pretrained generator’s image manifold and therefore relies more strongly on additional samples to cover diverse patterns.

\begin{table}[t]
\centering
\caption{
Comparison of downstream validation accuracy between SRG and baseline methods on fine-grained ImageNet subsets under different IPC settings using DINOv2. The best result in each setting is highlighted in bold.
}
\label{tab:subset}
\begin{tabular}{llccc}
\toprule
Subset & Method & IPC=1 & IPC=3 & IPC=5 \\
 
\midrule
\multirow{3}{*}{Woof} 
& DiT & $83.7_{\pm 1.7}$ & $89.1_{\pm 0.4}$ & $88.7_{\pm 0.6}$ \\
& MGD3 & $78.6_{\pm 1.1}$ & $89.0_{\pm 0.3}$ & $89.3_{\pm 0.6}$ \\
& LGM & \bm{$88.5_{\pm 0.8}$} & $90.1_{\pm 0.3}$ & $90.8_{\pm 0.3}$ \\
& SRG & $88.4_{\pm 0.7}$ & \bm{$91.4_{\pm 0.1}$} & \bm{$92.0_{\pm 0.5}$} \\

\midrule
\multirow{4}{*}{Fruits}
& DiT  & $77.7_{\pm 2.1}$ & $82.2_{\pm 0.6}$ & $84.3_{\pm 0.5}$ \\
& MGD3 & $77.4_{\pm 0.6}$ & $82.4_{\pm 1.1}$ & $85.4_{\pm 1.0}$ \\
& LGM  & \bm{$86.6_{\pm 0.8}$} & $87.4_{\pm 0.9}$ & $87.5_{\pm 0.6}$ \\
& SRG  & $85.4_{\pm 0.9}$ & \bm{$88.1_{\pm 0.4}$} & \bm{$89.1_{\pm 0.7}$} \\


\midrule
\multirow{3}{*}{\makecell{Instru- \\ ments}}
& DiT  & $64.7_{\pm 1.5}$ & $65.9_{\pm 0.4}$ & $76.3_{\pm 0.9}$ \\
& MGD3 & $49.9_{\pm 2.0}$ & $67.8_{\pm 0.8}$ & $73.5_{\pm 1.0}$ \\
& LGM  & \bm{$70.6_{\pm 1.5}$}  & \bm{$81.4_{\pm 0.4}$}  & $82.3_{\pm 0.1}$ \\
& SRG  & $70.2_{\pm 1.4}$ & $81.0_{\pm 0.6}$ & \bm{$82.6_{\pm 0.2}$} \\

\bottomrule
\end{tabular}
\end{table}

\subsection{Fine-Grained Classification}
We further evaluate SRG in fine-grained classification settings, where successful distillation requires preserving subtle
inter-class differences. Specifically, we compare the performance of SRG with baseline methods on three ImageNet-1K subsets containing visually similar classes, namely \textit{Woof}, \textit{Fruits}, and \textit{Instruments}. As shown in Table~\ref{tab:subset}, SRG outperforms the generative baselines in all the settings and outperforms LGM in most high-IPC settings. These results support the effectiveness of SSL-space guidance for datasets whose classes
have similar visual content.

\begin{table}[t]
\centering
\small
\caption{
Cross-encoder evaluation of downstream validation accuracy on ImageNet-100 under $\mathrm{IPC} = 5$. The Average row reports the mean and standard deviation across the four evaluation encoders.
}
\label{tab:cross_model}
\begin{tabular}{p{30pt}C{24pt}C{24pt}C{30pt}C{30pt}C{24pt}}
\toprule
\multirow{2}{*}{\makecell[l]{Evaluation \\ Encoder}}
& \multicolumn{4}{c}{Distillation Encoder}
& \multirow{2}{*}{Full}
\\

\cmidrule{2-5}
& CLIP & DINOv2 & EVA-02 & MoCov3 & 
\\

\midrule
CLIP 
& $87.8_{\pm 0.1}$ 
& $85.0_{\pm 0.0}$ 
& $85.0_{\pm 0.1}$ 
& $84.6_{\pm 0.1}$ 
& $92.5_{\pm 0.0}$ 
\\

DINOv2
& $91.1_{\pm 0.1}$ 
& $93.2_{\pm 0.1}$ 
& $90.6_{\pm 0.1}$ 
& $90.8_{\pm 0.1}$ 
& $95.2_{\pm 0.1}$ 
\\

EVA-02 
& $88.6_{\pm 0.1}$ 
& $88.4_{\pm 0.1}$ 
& $90.4_{\pm 0.1}$ 
& $88.7_{\pm 0.1}$ 
& $94.1_{\pm 0.1}$ 
\\

MoCov3
& $84.0_{\pm 0.1}$ 
& $83.3_{\pm 0.1}$ 
& $82.6_{\pm 0.0}$ 
& $84.8_{\pm 0.1}$ 
& $89.4_{\pm 0.3}$ 
\\

\midrule
Average
& $87.9_{\pm 2.6}$ 
& $87.5_{\pm 3.8}$ 
& $87.2_{\pm 2.9}$ 
& $87.2_{\pm 2.3}$ 
& $92.8_{\pm 2.2}$ 
\\

\bottomrule
\end{tabular}
\end{table}

\begin{figure}[t]
    \centering
    \includegraphics[width=\linewidth]{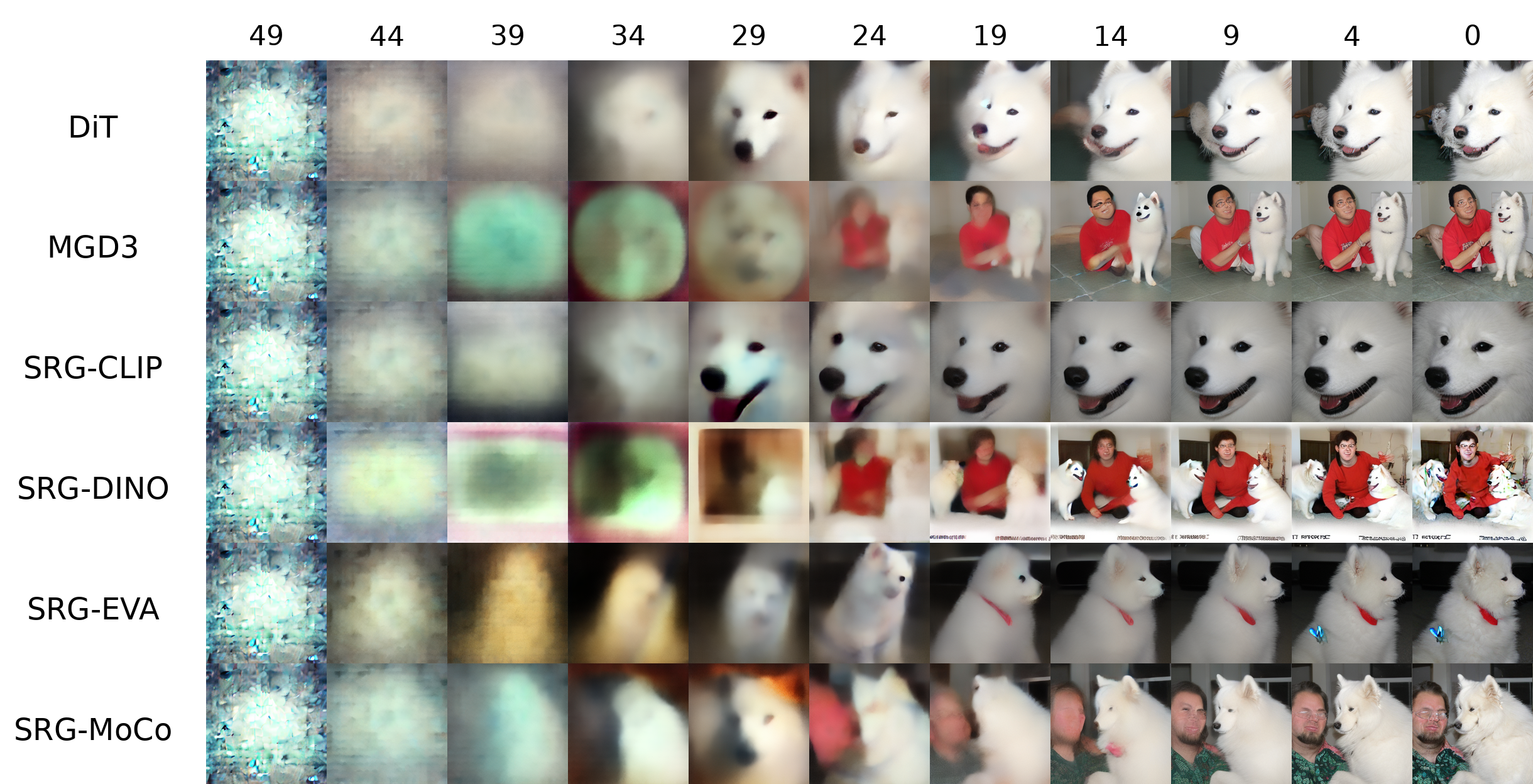}
    \caption{Comparison of intermediate images at different denoising timesteps for DiT, MGD3, and SRG. The denoising timesteps are arranged from $49$ to $0$.}
    \label{fig:denoise}
\end{figure}
\begin{figure}[t]
    \centering
    \includegraphics[width=\linewidth]{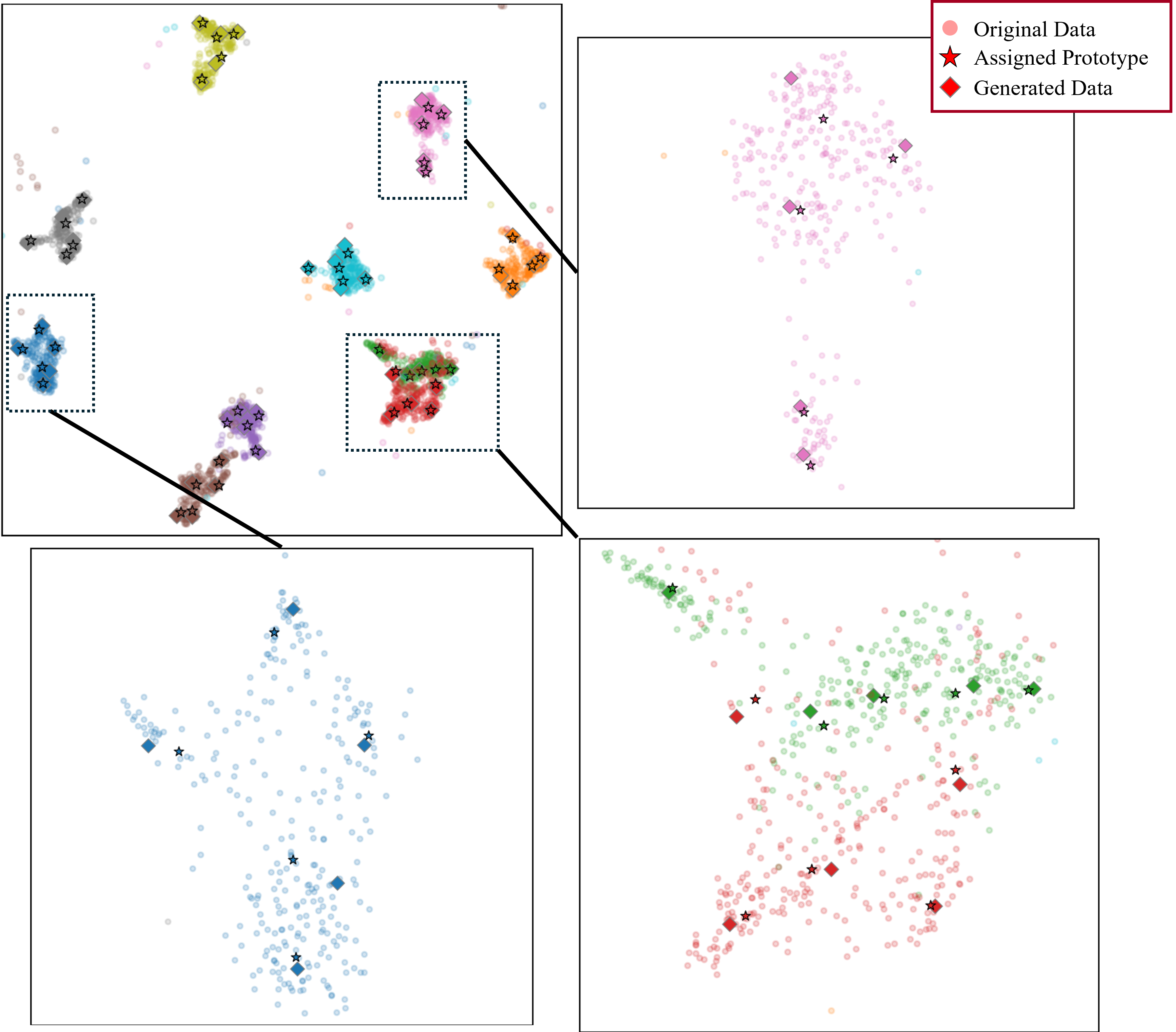}
    \caption{ 
    UMAP visualization of real samples, class prototypes, and SRG-generated samples in the SSL representation space on ImageNet-Woof at $\mathrm{IPC} = 5$.
    }
    \label{fig:umap}
\end{figure}
\subsection{Cross-Encoder Generalization}
To examine the generalization of SRG across pretrained representation spaces, we conduct a cross-encoder evaluation on ImageNet-100 at IPC $= 5$. We consider four SSL encoders: CLIP \cite{radford2021clip}, DINOv2 \cite{oquab2024dinov2}, EVA-02 \cite{fang2024eva02}, and MoCov3 \cite{chen2021mocov3}. Each encoder is used separately to guide distillation, and every resulting distilled dataset is then evaluated with all four encoders. This protocol measures both SRG’s performance with different distillation encoders and its transfer to representation spaces not used during distillation. For each distillation encoder, we also report the mean and standard deviation of accuracy across the four evaluation encoders.

\par

As shown in Table~\ref{tab:cross_model}, accuracy varies across evaluation encoders even when the full dataset is used, reflecting inherent differences in their representation quality. For each evaluation encoder, the highest accuracy occurs when the same encoder guides distillation. The mean accuracies of datasets distilled with the four encoders lie within a narrow range, indicating stable aggregate performance across distillation encoders. Additional experiments across different IPC settings and comparisons with LGM are provided in the supplementary material.

\subsection{Visualization}
We provide two qualitative analyses of SRG’s behavior. First, Fig.~\ref{fig:denoise} shows intermediate images decoded from the predicted clean latent at different denoising timesteps for DiT, MGD3, and SRG on class n02111889 (Samoyed) using the same initial noise. Under SRG’s schedule, latent-space guidance is active at early stages ($t_i \geq 20$), when coarse image structure emerges, whereas SSL-space guidance is active at later stages ($t_i < 20$), when semantic details become visible. The trajectories and final appearances also vary among guidance encoders, indicating different representational preferences across the encoders.

\par

Second, we use UMAP \cite{McInnes2018umap} to visualize the distributions of real samples, assigned prototypes, and generated samples in the SSL representation space on ImageNet-Woof. As shown in Fig.~\ref{fig:umap}, SRG-generated samples generally remain near their assigned prototypes. In the displayed well-separated class, generated samples occupy peripheral portions of the class cluster. In comparison, for the overlapping classes, several generated samples lie on the sides of their clusters farther from neighboring classes. These qualitative patterns are consistent with the intended objectives of prototype alignment, intra-class assignment, and inter-class discrimination. Additional visualization analyses are provided in the supplementary material.

\begin{figure}
\begin{subfigure}[t]{0.48\linewidth}
    \centering
    \includegraphics[width=\linewidth]{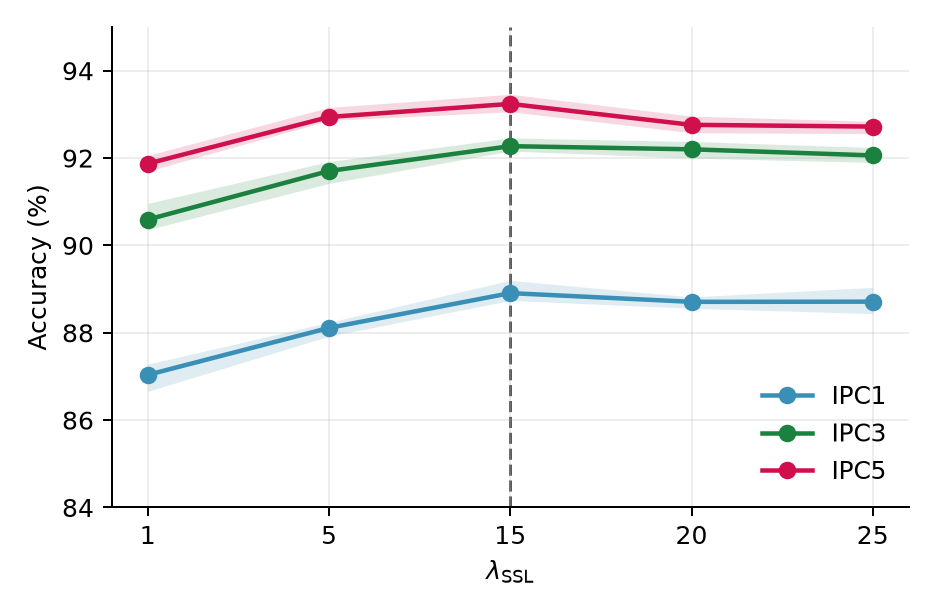}
    \caption{Guidance scale $\lambda_{\mathrm{SSL}}$.}
    \label{fig:ssl_scale}
\end{subfigure}
\hfill
\begin{subfigure}[t]{0.48\linewidth}
    \centering
    \includegraphics[width=\linewidth]{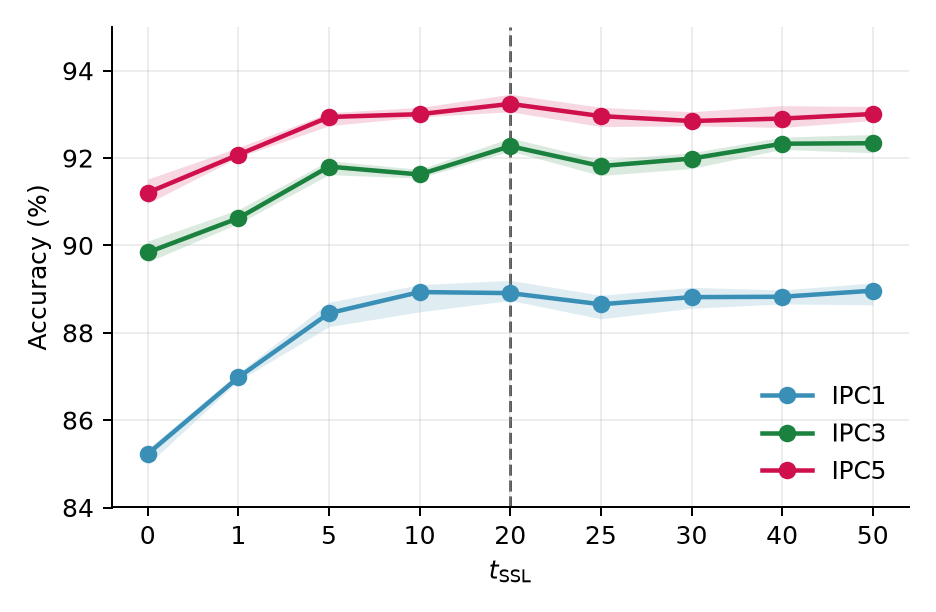}
    \caption{Starting timestep $t_{\mathrm{SSL}}$.}
    \label{fig:ssl_timestep}
\end{subfigure}
\caption{
Sensitivity analysis of the SSL-space guidance scale $\lambda_{\mathrm{SSL}}$ and starting timestep $t_{\mathrm{SSL}}$ on ImageNet-100 under different IPC settings. Points denote mean accuracy, and shaded regions indicate the minimum-to-maximum range.
}
\end{figure}

\begin{table}[t]
\centering
\caption{
Ablation study of latent-space guidance $\bm{g}_{lat}^i$ and SSL-space guidance $ \bm{g}_{\SSL}^i$ on ImageNet-100 under different IPC settings. The best result in each setting is highlighted in bold.
}
\label{tab:ablation}
\begin{tabular}{cc|ccc}
\toprule
$\bm{g}_{lat}^i$ 
& $\bm{g}_{\SSL}^i$ 
& IPC=1 & IPC=3 & IPC=5 \\
\midrule
\xmark & \xmark
& $83.3_{\pm 0.2}$ & $87.9_{\pm 0.2}$ & $89.1_{\pm 0.1}$ 
\\

\xmark & \cmark 
& $88.9_{\pm 0.1}$ & $91.2_{\pm 0.2}$ & $92.3_{\pm 0.2}$ 
\\

\cmark & \xmark 
& $85.4_{\pm 0.2}$ & $89.8_{\pm 0.1}$ & $91.3_{\pm 0.1}$  
\\

\cmark & \cmark 
& \bm{$89.0_{\pm 0.1}$} & \bm{$92.3_{\pm 0.1}$} & \bm{$93.2_{\pm 0.2}$} 
\\

\bottomrule
\end{tabular}
\end{table}

\subsection{Ablation Study}
We conduct an ablation study on ImageNet-100 to evaluate the contributions of latent-space guidance and SSL-space guidance.  
As shown in Table~\ref{tab:ablation}, latent-space guidance alone brings a modest improvement over the unguided baseline, while SSL-space guidance yields more substantial gains. Combining them consistently achieves the best accuracy across all IPC settings, supporting the complementarity of the two signals. Detailed ablations of the individual SSL-space objectives are provided in the supplementary material.

\subsection{Sensitivity to Hyperparameters}
We further analyze the sensitivity of SRG to two hyperparameters associated with SSL-space guidance: the guidance scale $\lambda_{\mathrm{SSL}}$ and the starting timestep $t_{\mathrm{SSL}}$. Analyses of the remaining hyperparameters are provided in the supplementary material. As shown in Fig.~\ref{fig:ssl_scale}, performance is relatively stable across the tested intermediate values of $\lambda_{\mathrm{SSL}}$. A very small $\lambda_{\mathrm{SSL}}$ leads to lower accuracy, likely because it provides insufficient representation-level supervision. Increasing $\lambda_{\mathrm{SSL}}$ beyond an intermediate range yields no further improvement, possibly because overly strong guidance interferes with DiT's original denoising trajectory. Among the evaluated settings, $\lambda_{\mathrm{SSL}}=15$ achieves the best overall performance across IPC budgets. Fig.~\ref{fig:ssl_timestep} shows that applying SSL-space guidance only during the final several denoising steps already substantially improves performance over using no SSL-space guidance. Extending the guidance to earlier timesteps provides only limited additional gains while increasing the computational cost. We therefore set $t_{\mathrm{SSL}}=20$ to balance performance and computational efficiency.

\begin{table}[t]
\centering
\small
\caption{
Comparison of runtime between SRG and baseline methods on different ImageNet benchmarks under various IPC settings, with the prototype-construction cost of SRG reported separately. All runtimes are reported in minutes.
}
\label{tab:time}
\begin{threeparttable}
\begin{tabular}{cccccc}
\toprule
\multirow{2}{*}{Dataset}
& \multirow{2}{*}{Method}
& \multicolumn{2}{c}{Time (Min)} 
& \multicolumn{2}{c}{Mem. (GB)}
\\

\cmidrule(lr){3-4} \cmidrule(lr){5-6}
& & 1 & 3 & 1 & 3
\\

\midrule
\multirow{4}{*}{\makecell[c]{ImageNet- \\ IDC}}
& LGM
& 34.8 & 75.1
& 10.2 & 28.8
\\

& CLP-DD
& 1.4 & 2.5
& 1.1 & 2.3
\\

& DiT
& 0.2 & 0.5
& 3.2 & 3.2
\\

& SRG
& 0.3 & 0.9
& 5.8 & 5.8
\\

& Prototype
& 0.5 & 0.6
& 9.5 & 9.5
\\

\midrule
\multirow{4}{*}{\makecell[c]{ImageNet- \\ 100}}
& LGM
& 81.8 & OOM
& 28.7 & > 48
\\

& CLP-DD
& 8.7 & 26.8
& 6.4 & 18.3
\\

& DiT
& 1.6 & 4.7
& 3.2 & 3.2
\\

& SRG
& 2.9 & 8.5
& 5.8 & 5.8
\\

& Prototype
& 2.7 & 2.7
& 9.5 & 9.5
\\

\midrule
\multirow{4}{*}{\makecell[c]{ImageNet- \\ 1K}}
& LGM
& OOM\tnote{*} & OOM
& > 48 & > 48
\\

& CLP-DD
& OOM & OOM
& > 48 & > 48
\\

& DiT
& 15.5 & 46.4
& 3.2 & 3.2
\\

& SRG
& 27.8 & 83.3
& 5.8 & 5.8
\\

& Prototype
& 36.8 & 39.3
& 9.5 & 9.5
\\

\bottomrule
\end{tabular}

\begin{tablenotes}
\footnotesize
\item[*] The original paper reports a runtime of approximately 720 minutes using four H200 GPUs.
\end{tablenotes}

\end{threeparttable}
\end{table}

\subsection{Computational Cost}
\label{sec:cost}
Table~\ref{tab:time} compares the computational costs of LGM, CLP-DD, DiT, and SRG on a single NVIDIA A6000 GPU, with the prototype-construction cost reported separately. SRG introduces moderate sampling-time and memory overhead over DiT due to SSL-space guidance, while remaining more efficient than the evaluated optimization-based baselines. Prototype construction requires one-time preprocessing, after which the extracted representations are cached for reuse. CLP-DD likewise requires representation extraction as part of its preprocessing. Under this hardware setting, SRG exhibits favorable scalability to large-scale datasets while maintaining moderate overhead over the generative baseline.

\section{Conclusion}
In this paper, we proposed SRG, a generative dataset distillation framework for downstream adaptation with pretrained SSL encoders. SRG combines stage-wise latent-space anchoring with SSL-space objectives for prototype alignment, inter-class discrimination, and intra-class assignment. Experiments across multiple datasets and IPC settings demonstrate consistent improvements over generative baselines, competitive performance against optimization-based methods, and favorable computational scalability. Cross-encoder evaluation further indicates transfer across pretrained representation spaces. Nevertheless, its reliance on an ImageNet-pretrained generator currently limits its applicability to broader visual domains. 

\bibliography{aaai2027}

\newpage

\section{Algorithm}
We provide detailed pseudocode for SRG using the notation and equation numbers from the main paper. As shown in Algorithm~\ref{alg:srg}, SRG first constructs class-specific SSL prototypes and retrieves the nearest real sample to each prototype as the latent-guidance target. It then generates one synthetic sample for each prototype through a stage-wise guided denoising process, in which latent-space guidance is applied during the early stages, and SSL-space guidance is applied during the later stages. Finally, the generated samples from all classes are collected to form the distilled dataset.

\begin{flushleft}
\rule{\linewidth}{0.6pt}
\refstepcounter{algorithm}
Algorithm \thealgorithm: Self-Supervised Representation-Guided Generative Dataset Distillation (SRG).
\label{alg:srg}
\rule{\linewidth}{0.4pt}
\end{flushleft}
{\footnotesize \itshape Equation numbers refer to the main paper.}

\begin{algorithmic}[1]
\small
\REQUIRE
$\mathcal{D}=\{(\bm{x}_i,y_i)\}_{i=1}^{N}$: original dataset;
$f_{\theta}$: pretrained DiT;
$f_{\phi}$: pretrained SSL encoder;
$E_{\psi},D_{\psi}$: VAE encoder and decoder~\cite{kingma2013VAE};
$K$: IPC;
$\mathcal{T}=(t_0,\ldots,t_{T-1})$: decreasing denoising schedule;
$t_{\mathrm{lat}},t_{\mathrm{SSL}},
\lambda_{\mathrm{lat}},\lambda_{\mathrm{SSL}},
\tau_{\mathrm{inter}},\tau_{\mathrm{intra}}$:
guidance hyperparameters.
\ENSURE
$\mathcal{S}$: distilled dataset.

\STATE Initialize $\mathcal{S}\leftarrow\varnothing$
\STATE \textbf{Stage 1: Prototype and latent-anchor construction}
\FOR{each class $c=1,\ldots,C$}
    \STATE Collect $\mathcal{D}_c=\{\bm{x}_n^c\}_{n=1}^{N_c}$ and compute
    $\bm{h}_n^c\leftarrow\operatorname{norm}_2(f_{\phi}(\bm{x}_n^c))$ for all $n$
    \STATE Obtain $\{\bm{p}_{c,k}\}_{k=1}^{K}$ by spherical $K$-means with Eq.~(4)
    \FOR{$k=1,\ldots,K$}
        \STATE $n_{c,k}^{\star}\leftarrow\arg\max_{1\leq n\leq N_c}
        \operatorname{sim}(\bm{h}_n^c,\bm{p}_{c,k})$
        \STATE $\bm{a}_{c,k}\leftarrow E_{\psi}(\bm{x}_{n_{c,k}^{\star}}^c)$
    \ENDFOR
\ENDFOR

\STATE \textbf{Stage 2: Stage-wise guided generation}
\FOR{each class $c=1,\ldots,C$}
    \FOR{each prototype $k=1,\ldots,K$}
        \STATE Sample $\bm{v}_0\sim\mathcal{N}(\bm{0},\bm{I})$
        \FOR{$i=0,\ldots,T-1$}
            \STATE Set $t_i\leftarrow\mathcal{T}[i]$ and obtain
            $\mu_{\theta}(\bm{v}_i,t_i,c)$ and $\hat{\bm{z}}_0^i$ from DiT
            \STATE Initialize $\bm{g}_{\mathrm{lat}}^i,\bm{g}_{\mathrm{SSL}}^i\leftarrow\bm{0}$
            \IF{$t_i>t_{\mathrm{lat}}$}
                \STATE Compute $\bm{g}_{\mathrm{lat}}^i$ with target $\bm{m}_{c,k}$ with Eq.~(2)
            \ENDIF
            \IF{$t_i<t_{\mathrm{SSL}}$}
                \STATE $\bm{r}_i\leftarrow\operatorname{norm}_2\!\left(
                f_{\phi}\!\left(D_{\psi}(\hat{\bm{z}}_0^i)\right)\right)$
                \STATE Compute $\mathcal{L}_{\mathrm{proto}}^i$, $\mathcal{L}_{\mathrm{inter}}^i$, and
                $\mathcal{L}_{\mathrm{intra}}^i$ with Eqs.~(6)--(8)
                \STATE Compute $\mathcal{L}_{\mathrm{SSL}}^i$ and $\bm{g}_{\mathrm{SSL}}^i$
                with Eqs.~(9) and (10)
            \ENDIF
            \STATE Compute $\bm{g}_{\mathrm{SRG}}^i$ with Eq.~(11)
            \STATE Sample $\bm{\epsilon}_i\sim\mathcal{N}(\bm{0},\bm{I})$ and update
            $\bm{v}_{i+1}$ with Eq.~(12)
        \ENDFOR
        \STATE $\tilde{\bm{x}}_{c,k}\leftarrow D_{\psi}(\bm{v}_T)$ and
        $\mathcal{S}\leftarrow\mathcal{S}\cup\{(\tilde{\bm{x}}_{c,k},c)\}$
    \ENDFOR
\ENDFOR
\RETURN $\mathcal{S}$
\end{algorithmic}
\noindent\rule{\linewidth}{0.4pt}

\section{Implementation Details}
We use four pretrained SSL encoders with ViT-B backbones~\cite{dosovitskiy2021vit}: CLIP ViT-B/16~\cite{radford2021clip}, DINOv2 ViT-B/14~\cite{oquab2024dinov2}, EVA-02 ViT-B/14~\cite{fang2024eva02}, and MoCo-v3 ViT-B~\cite{chen2021mocov3}. The feature dimension is 512 for CLIP ViT-B/16 and 768 for DINOv2, EVA-02, and MoCo-v3. SSL features are extracted at a resolution of $224 \times 224$. 

\par

For SRG, we use a DiT-XL/2 backbone~\cite{Peebbles2023DiT} at an image resolution of $256 \times 256$, with $50$ denoising steps and a classifier-free guidance~\cite{ho2022clsFreeDiff} scale of $4.0$. Prototype centers are computed using spherical $K$-means in the normalized SSL feature space, with a maximum of $50$ iterations. The default parameters are the latent-guidance stopping timestep $t_{\mathrm{lat}}=20$, SSL-guidance starting timestep $t_{\mathrm{SSL}}=20$, SSL-guidance scale $\lambda_{\mathrm{SSL}}=15$, latent-guidance scale $\lambda_{\mathrm{lat}}=0.1$, inter-class discrimination temperature $\tau_{\mathrm{inter}}=1.0$, and intra-class assignment temperature $\tau_{\mathrm{intra}}=5.0$.

\par

Unless otherwise specified, the methods reproduced in our experiments use the default parameters provided by their original authors. For LGM~\cite{cazenavette2025lgm}, the number of augmentation views is set to $3$ on ImageNet-100~\cite{kim2022IDC} because of GPU memory limitations. For CLP-DD~\cite{peng2026clpdd}, we use the unpublished version code provided by the authors.

\begin{table*}[t]
\centering
\caption{
Cross-encoder comparison of downstream validation accuracy between SRG and LGM on ImageNet-100 at $\mathrm{IPC}=1$ and $\mathrm{IPC}=3$. Each average is computed from the mean accuracies obtained with the four evaluation encoders. The best average values are highlighted in bold for each distillation setting.
}
\label{sup:cross}
\setlength{\tabcolsep}{3.2pt}
\begin{tabular}{llcccccccc}
\toprule
& \multirow{2}{*}{\makecell[l]{Evaluation \\ Encoder}}
& \multicolumn{2}{c}{CLIP} 
& \multicolumn{2}{c}{DINOv2}
& \multicolumn{2}{c}{EVA-02}
& \multicolumn{2}{c}{MoCo-v3} 
\\

\cmidrule(lr){3-4} \cmidrule(lr){5-6} \cmidrule(lr){7-8} \cmidrule(lr){9-10}

&& LGM & SRG & LGM & SRG & LGM & SRG & LGM & SRG
\\

\midrule
\multirow{5}{*}{\makecell[l]{IPC = 1}}
& CLIP 
& $82.4_{\pm 0.1}$
& $78.5_{\pm 0.2}$
& $76.3_{\pm 0.2}$
& $71.6_{\pm 0.2}$
& $74.3_{\pm 0.2}$
& $74.5_{\pm 0.2}$
& $64.5_{\pm 0.2}$
& $72.7_{\pm 0.2}$
\\

& DINOv2
& $78.2_{\pm 0.3}$
& $85.2_{\pm 0.3}$
& $90.7_{\pm 0.1}$
& $88.7_{\pm 0.2}$
& $85.8_{\pm 0.3}$
& $85.2_{\pm 0.2}$
& $85.6_{\pm 0.1}$
& $84.0_{\pm 0.2}$
\\

& EVA-02 
& $80.8_{\pm 0.4}$
& $80.9_{\pm 0.4}$
& $86.0_{\pm 0.3}$
& $80.8_{\pm 0.2}$
& $88.5_{\pm 0.1}$
& $85.8_{\pm 0.2}$
& $81.2_{\pm 0.1}$
& $80.9_{\pm 0.1}$
\\

& MoCo-v3
& $55.7_{\pm 0.9}$
& $75.1_{\pm 0.0}$
& $76.3_{\pm 0.4}$
& $74.9_{\pm 0.1}$
& $64.8_{\pm 0.7}$
& $74.6_{\pm 0.1}$
& $83.1_{\pm 0.1}$
& $77.8_{\pm 0.1}$
\\

& Average
& $74.3_{\pm 10.8}$
& $\mathbf{79.9_{\pm 3.7}}$
& $\mathbf{82.3_{\pm 6.3}}$ 
& $79.0_{\pm 6.5}$
& $78.3_{\pm 9.4}$
& $\mathbf{80.0_{\pm 5.5}}$
& $78.6_{\pm 8.3}$
& $\mathbf{78.8_{\pm 4.2}}$
\\

\midrule
\multirow{5}{*}{\makecell[l]{IPC = 3}}
& CLIP 
& $86.5_{\pm 0.1}$
& $86.2_{\pm 0.1}$
& $80.4_{\pm 0.2}$
& $82.5_{\pm 0.1}$
& $79.7_{\pm 0.2}$
& $83.7_{\pm 0.1}$
& $71.3_{\pm 0.2}$
& $84.5_{\pm 0.1}$
\\

& DINOv2
& $83.0_{\pm 0.2}$ 
& $90.4_{\pm 0.2}$
& $91.7_{\pm 0.2}$
& $92.1_{\pm 0.0}$
& $88.0_{\pm 0.2}$
& $90.1_{\pm 0.2}$
& $86.4_{\pm 0.3}$
& $90.3_{\pm 0.2}$
\\

& EVA-02 
& $83.9_{\pm 0.2}$
& $87.0_{\pm 0.2}$
& $87.0_{\pm 0.3}$
& $88.0_{\pm 0.1}$
& $88.7_{\pm 0.1}$
& $89.6_{\pm 0.1}$
& $81.7_{\pm 0.4}$
& $88.3_{\pm 0.2}$
\\

& MoCo-v3
& $66.2_{\pm 0.6}$
& $82.1_{\pm 0.0}$
& $79.4_{\pm 0.5}$
& $81.8_{\pm 0.1}$
& $69.6_{\pm 0.3}$
& $80.9_{\pm 0.1}$
& $83.7_{\pm 0.0}$
& $84.5_{\pm 0.1}$
\\

& Average
& $79.9_{\pm 8.0}$ 
& $\mathbf{86.4_{\pm 3.0}}$
& $84.6_{\pm 5.0}$ 
& $\mathbf{86.1_{\pm 4.2}}$
& $81.5_{\pm 7.7}$ 
& $\mathbf{86.1_{\pm 3.9}}$
& $80.8_{\pm 5.7}$ 
& $\mathbf{86.9_{\pm 2.5}}$
\\

\bottomrule
\end{tabular}
\end{table*}

\par

For evaluation, we train a linear probe on the frozen SSL features for $1{,}000$ epochs and repeat each experiment five times. When augmenting the distilled training data for evaluation, we use resize-only preprocessing for CLP-DD and spatial augmentation for LGM, following the respective default settings provided by the authors. For the generative methods, we use weak photometric augmentation to introduce mild natural variation.

\par

All experiments are conducted on a GPU server equipped with an AMD EPYC 7413 24-core processor and four NVIDIA A6000 GPUs with 48 GB of memory each, running Ubuntu 22.04.4 LTS. Except for LGM on ImageNet-100 at IPC $=3$, which uses all four GPUs, each experiment uses a single A6000 GPU. The software environment consists of Python 3.14, PyTorch 2.12.0, torchvision 0.27.0, CUDA 12.6, and cuDNN 9.10. We use a default fixed random seed $0$. However, we do not enforce exact bitwise reproducibility because some optimized CUDA kernels used for efficient sampling are nondeterministic. Enabling fully deterministic algorithms would substantially slow sampling and may disable some operations. 

\section{Dataset Details}
For ImageNet-IDC and ImageNet-100, we follow the dataset configurations introduced by \citet{kim2022IDC}. ImageNet-100 is a subset of ImageNet-1K~\cite{deng2009imageNet,deng2015imageNet1k} containing 100 selected classes, while ImageNet-IDC contains 10 classes from ImageNet-100. The class composition of ImageNet-IDC is provided below. Each class is reported in the format ``ImageNet class index: class name (WordNet ID)'', where the class name corresponds to the first entry in the ImageNet label list.

\paragraph{ImageNet-IDC.}
452: Bonnet (n02869837);
64: Green Mamba (n01749939);
374: Langur (n02488291);
236: Doberman (n02107142);
993: Gyromitra (n13037406);
176: Saluki (n02091831);
882: Vacuum (n04517823);
904: Window Screen (n04589890);
503: Cocktail Shaker (n03062245);
74: Garden Spider (n01773797).

\par

For fine-grained evaluation, the \textit{Woof} subset follows the ImageWoof configuration of \citet{fastai2019imageNette}. We additionally construct the \textit{Fruits} and \textit{Instruments} subsets by selecting semantically related and visually similar classes from ImageNet-1K. Their class compositions are listed below.

\paragraph{\textit{Woof}.}
155: Shih-Tzu (n02086240);
159: Rhodesian Ridgeback (n02087394);
162: Beagle (n02088364);
167: English Foxhound (n02089973);
182: Border Terrier (n02093754);
193: Australian Terrier (n02096294);
207: Golden Retriever (n02099601);
229: Old English Sheepdog (n02105641);
258: Samoyed (n02111889);
273: Dingo (n02115641).

\paragraph{\textit{Fruits}.}
953: Pineapple (n07753275);
954: Banana (n07753592);
949: Strawberry (n07745940);
950: Orange (n07747607);
951: Lemon (n07749582);
957: Pomegranate (n07768694);
952: Fig (n07753113);
945: Bell Pepper (n07720875);
943: Cucumber (n07718472);
948: Granny Smith (n07742313).

\paragraph{\textit{Instruments}.}
432: Bassoon (n02804610);
513: Cornet (n03110669);
558: Flute (n03372029);
566: French Horn (n03394916);
593: Harmonica (n03494278);
683: Oboe (n03838899);
684: Ocarina (n03840681);
699: Panpipe (n03884397);
776: Saxophone (n04141076);
875: Trombone (n04487394).

\begin{table*}[!]
\centering
\caption{
Joint sensitivity analysis of the latent-guidance stopping timestep $t_{\mathrm{lat}}$ and SSL-guidance starting timestep $t_{\mathrm{SSL}}$ on ImageNet-100 at
IPC $=5$. The default setting is underlined, and the best result is highlighted in bold.}
\label{sup:time_step}
\begin{tabular}{c|ccccccc}
\toprule
$t_{\mathrm{lat}} \backslash t_{\mathrm{SSL}}$
& 0 & 10 & 20 & 25 & 30 & 40 & 50 
\\
\midrule

51
& $89.12_{\pm 0.06}$
& $91.79_{\pm 0.11}$
& $92.29_{\pm 0.15}$
& $92.02_{\pm 0.15}$
& $92.28_{\pm 0.17}$
& $92.16_{\pm 0.07}$
& $92.24_{\pm 0.09}$
\\

40
& $89.28_{\pm 0.12}$
& $91.58_{\pm 0.14}$
& $91.79_{\pm 0.15}$
& $92.24_{\pm 0.15}$
& $92.53_{\pm 0.15}$
& $92.01_{\pm 0.15}$
& $92.31_{\pm 0.15}$
\\

30
& $89.88_{\pm 0.15}$
& $92.06_{\pm 0.16}$
& $92.77_{\pm 0.09}$
& $92.71_{\pm 0.08}$
& $92.70_{\pm 0.08}$
& $92.64_{\pm 0.15}$
& $92.52_{\pm 0.15}$
\\

25
& $90.51_{\pm 0.15}$
& $92.54_{\pm 0.10}$
& $92.93_{\pm 0.18}$
& $92.80_{\pm 0.08}$
& $93.03_{\pm 0.15}$
& $92.89_{\pm 0.12}$
& $93.06_{\pm 0.09}$
\\

20
& $91.21_{\pm 0.19}$
& $93.00_{\pm 0.08}$
& $\underline{\mathbf{93.24_{\pm 0.15}}}$
& $92.96_{\pm 0.15}$
& $92.85_{\pm 0.14}$
& $92.90_{\pm 0.18}$
& $93.01_{\pm 0.12}$ 
\\

10
& $91.28_{\pm 0.10}$
& $92.68_{\pm 0.14}$
& $93.03_{\pm 0.10}$
& $92.80_{\pm 0.07}$
& $93.02_{\pm 0.15}$
& $92.83_{\pm 0.05}$ 
& $93.11_{\pm 0.13}$ 
\\

0
& $91.33_{\pm 0.09}$
& $92.49_{\pm 0.10}$
& $92.74_{\pm 0.10}$
& $92.91_{\pm 0.14}$
& $92.96_{\pm 0.06}$
& $92.88_{\pm 0.14}$
& $93.03_{\pm 0.11}$ 
\\

\bottomrule
\end{tabular}
\end{table*}

\section{Cross-Encoder Generalization}
We further validate SRG in the cross-encoder setting on more IPC settings, and conduct comparison with LGM. As shown in Table~\ref{sup:cross}, at IPC $=1$, LGM generally retains an advantage when the same encoder is used for both distillation and evaluation. However, its performance transfers less consistently across encoders, with several combinations exhibiting substantial degradation and less stable average performance than SRG. At IPC $=3$, SRG outperforms LGM for nearly all encoder combinations and achieves substantially higher average performance, indicating better scalability and transfer across encoders. One possible explanation is that LGM directly optimizes synthetic data for a specific encoder, which can yield stronger encoder-specific adaptation but weaker transfer to other representation spaces. In contrast, SRG performs distillation by guiding a pretrained generative model. In addition to encoder-specific SSL guidance, the model's visual prior preserves general object structure and appearance, thereby reducing overfitting to the distillation encoder.

\section{Detailed Hyperparameter Analyses}
\subsection{Guidance Timesteps}
Table~\ref{sup:time_step} examines the joint effect of the latent-guidance stopping timestep $t_{\mathrm{lat}}$ and the SSL-guidance starting timestep $t_{\mathrm{SSL}}$ on ImageNet-100 at IPC $=5$. As the differences in the experimental results were not clear under some settings, all values are reported to two decimal places for clarity. At the boundary setting $t_{\mathrm{SSL}}=0$, where SSL-space guidance is disabled, extending latent-space guidance beyond the earliest denoising stages initially provides clear improvements, but the gains diminish at later timesteps. Similarly, at $t_{\mathrm{lat}}=51$, where latent-space guidance is disabled, introducing SSL-space guidance during the later denoising stages substantially improves performance, whereas extending it to earlier timesteps produces only limited additional gains, consistent with the discussion in the main paper. These observations agree with the intended roles of the two guidance signals: latent-space guidance mainly helps determine coarse structure during early denoising, whereas SSL-space guidance mainly helps refine semantic details during later stages. A similar trend is observed when the two guidance intervals overlap, suggesting that the two signals can operate together effectively. Because both guidance operations introduce additional computation, particularly the representation mapping and gradient backpropagation required by SSL-space guidance, we select $t_{\mathrm{lat}}=t_{\mathrm{SSL}}=20$ as the default setting to balance performance and computational cost. 

\par
\begin{table}[t]
\centering
\small
\caption{Joint sensitivity analysis of the temperature parameters $\tau_{\mathrm{inter}}$ and $\tau_{\mathrm{intra}}$ on ImageNet-100 at IPC $=5$. The default setting is underlined, and the best result is highlighted in bold.}
\label{sup:temperature}
\small
\setlength{\tabcolsep}{5pt}
\begin{tabular}{cccccc}
\toprule
\multirow[b]{2}{*}{$\tau_{\mathrm{inter}}$}
& \multicolumn{5}{c}{$\tau_{\mathrm{intra}}$}
\\
\cmidrule{2-6}
& 0.1 & 0.5 & 1.0 & 5.0 & 10.0 \\
\midrule
0.1 
& $91.2_{\pm 0.2}$ 
& $92.0_{\pm 0.1}$ 
& $92.2_{\pm 0.1}$ 
& $91.6_{\pm 0.2}$ 
& $91.7_{\pm 0.1}$
\\

0.5 
& $92.2_{\pm 0.1}$ 
& $92.5_{\pm 0.2}$ 
& $93.0_{\pm 0.1}$ 
& $92.7_{\pm 0.2}$ 
& $92.9_{\pm 0.2}$ 
\\

1.0 
& $91.7_{\pm 0.1}$ 
& $92.8_{\pm 0.1}$ 
& $93.0_{\pm 0.1}$ 
& $\underline{\mathbf{93.2_{\pm 0.2}}}$ 
& $93.1_{\pm 0.1}$
\\

5.0 
& $92.3_{\pm 0.1}$
& $92.5_{\pm 0.1}$
& $92.3_{\pm 0.1}$ 
& $92.8_{\pm 0.1}$ 
& $92.7_{\pm 0.1}$

\\

10.0
& $92.2_{\pm 0.1}$ 
& $92.7_{\pm 0.1}$ 
& $93.0_{\pm 0.1}$ 
& $92.7_{\pm 0.1}$ 
& $92.9_{\pm 0.2}$ 
\\

\bottomrule
\end{tabular}
\end{table}

\begin{table}[t]
\centering
\caption{Joint sensitivity analysis of the guidance scales $\lambda_{\mathrm{lat}}$ and $\lambda_{\SSL}$ on ImageNet-100 at IPC $=5$. The default setting is underlined, and the best result is highlighted in bold.}
\label{sup:scale}
\small
\setlength{\tabcolsep}{5pt}
\begin{tabular}{cccccc}
\toprule
\multirow[b]{2}{*}{$\lambda_{\mathrm{SSL}}$}
& \multicolumn{5}{c}{$\lambda_{\mathrm{lat}}$}
\\
\cmidrule{2-6}
& 0.01 & 0.05 & 0.1 & 0.2 & 0.3 \\
\midrule

1 
& $89.9_{\pm 0.2}$ 
& $90.8_{\pm 0.0}$ 
& $92.0_{\pm 0.0}$
& $92.3_{\pm 0.1}$ 
& $88.1_{\pm 0.3}$ 
\\


10
& $92.0_{\pm 0.2}$ 
& $92.7_{\pm 0.2}$ 
& $93.1_{\pm 0.1}$ 
& $92.8_{\pm 0.1}$ 
& $90.2_{\pm 0.1}$ 
\\

15 
& $92.4_{\pm 0.1}$ 
& $93.0_{\pm 0.0}$ 
& $\underline{\mathbf{93.2_{\pm 0.2}}}$ 
& $93.1_{\pm 0.1}$ 
& $90.4_{\pm 0.1}$ 
\\

20 
& $92.6_{\pm 0.1}$ 
& $92.6_{\pm 0.1}$ 
& $92.8_{\pm 0.1}$ 
& $93.0_{\pm 0.1}$ 
& $90.5_{\pm 0.1}$ 
\\

30
& $92.4_{\pm 0.1}$ 
& $92.6_{\pm 0.1}$ 
& $92.3_{\pm 0.1}$ 
& $92.6_{\pm 0.1}$ 
& $90.6_{\pm 0.2}$ 
\\

\bottomrule
\end{tabular}
\end{table}

\subsection{Temperature parameters}
The temperature parameters $\tau_{\mathrm{inter}}$ and $\tau_{\mathrm{intra}}$ control the sharpness of their corresponding objectives. A smaller temperature makes an objective more sensitive to the most similar prototype, whereas a larger temperature distributes weight more evenly across prototypes. As shown in Table~\ref{sup:temperature}, accuracy generally increases and then decreases as either temperature grows. We adopt $\tau_{\mathrm{inter}}=1.0$ and $\tau_{\mathrm{intra}}=5.0$ as the default values because this combination achieves the highest accuracy. The difference between the selected temperatures is consistent with the structures modeled by the two objectives. Prototypes from other classes are typically more separated from the predicted representation, making the nearest ones particularly informative for discrimination. In contrast, intra-class prototypes are distributed more closely around the predicted representation, and repelling the representation from only one competing prototype may move it closer to another.

\subsection{Guidance scales}
The guidance scales $\lambda_{\mathrm{lat}}$ and $\lambda_{\mathrm{SSL}}$ determine the strengths of latent-space and SSL-space guidance, respectively. When a scale is too small, the corresponding guidance has only a limited effect on the sampling trajectory, resulting in performance close to that of unguided DiT. Increasing either guidance scale produces clear improvements, whereas excessively strong guidance can dominate the diffusion and steer samples toward regions not supported by the pretrained generative prior, resulting in performance degradation. We set $\lambda_{\mathrm{lat}}=0.1$ and $\lambda_{\mathrm{SSL}}=15$ as the default values because this combination achieves the best performance.

\begin{table}[t]
\centering
\small
\caption{Joint sensitivity analysis of the objective weights $w_{\mathrm{inter}}$ and $w_{\mathrm{intra}}$ on ImageNet-100 at IPC $=5$. The default setting is underlined, and the best result is highlighted in bold.}
\label{sup:weight}
\setlength{\tabcolsep}{5pt}
\begin{tabular}{cccccc}
\toprule
\multirow[b]{2}{*}{$w_{\mathrm{inter}}$}
& \multicolumn{4}{c}{$w_{\mathrm{intra}}$}
\\
\cmidrule{2-6}
& 0.1 & 1.0 & 5.0 & 10.0 & 30.0 \\
\midrule

0.1 
& $92.9_{\pm 0.1}$ 
& $92.9_{\pm 0.2}$
& $92.8_{\pm 0.1}$ 
& $92.1_{\pm 0.1}$
& $89.6_{\pm 0.2}$ 
\\

1.0 
& $93.1_{\pm 0.1}$ 
& $\underline{\mathbf{93.2_{\pm 0.2}}}$
& $92.8_{\pm 0.1}$ 
& $92.4_{\pm 0.0}$
& $90.0_{\pm 0.2}$ 
\\

5.0 
& $93.1_{\pm 0.1}$ 
& $92.9_{\pm 0.1}$ 
& $92.9_{\pm 0.1}$ 
& $92.3_{\pm 0.1}$ 
& $90.0_{\pm 0.2}$ 
\\

10.0 
& $92.4_{\pm 0.1}$ 
& $92.7_{\pm 0.1}$ 
& $92.6_{\pm 0.3}$ 
& $92.4_{\pm 0.0}$ 
& $90.6_{\pm 0.2}$ 
\\

30.0 
& $89.3_{\pm 0.1}$ 
& $89.9_{\pm 0.1}$ 
& $88.6_{\pm 0.1}$
& $90.0_{\pm 0.1}$
& $87.4_{\pm 0.3}$ 
\\

\bottomrule
\end{tabular}
\end{table}

\begin{figure*}[t]
    \centering
    \includegraphics[width=\linewidth]{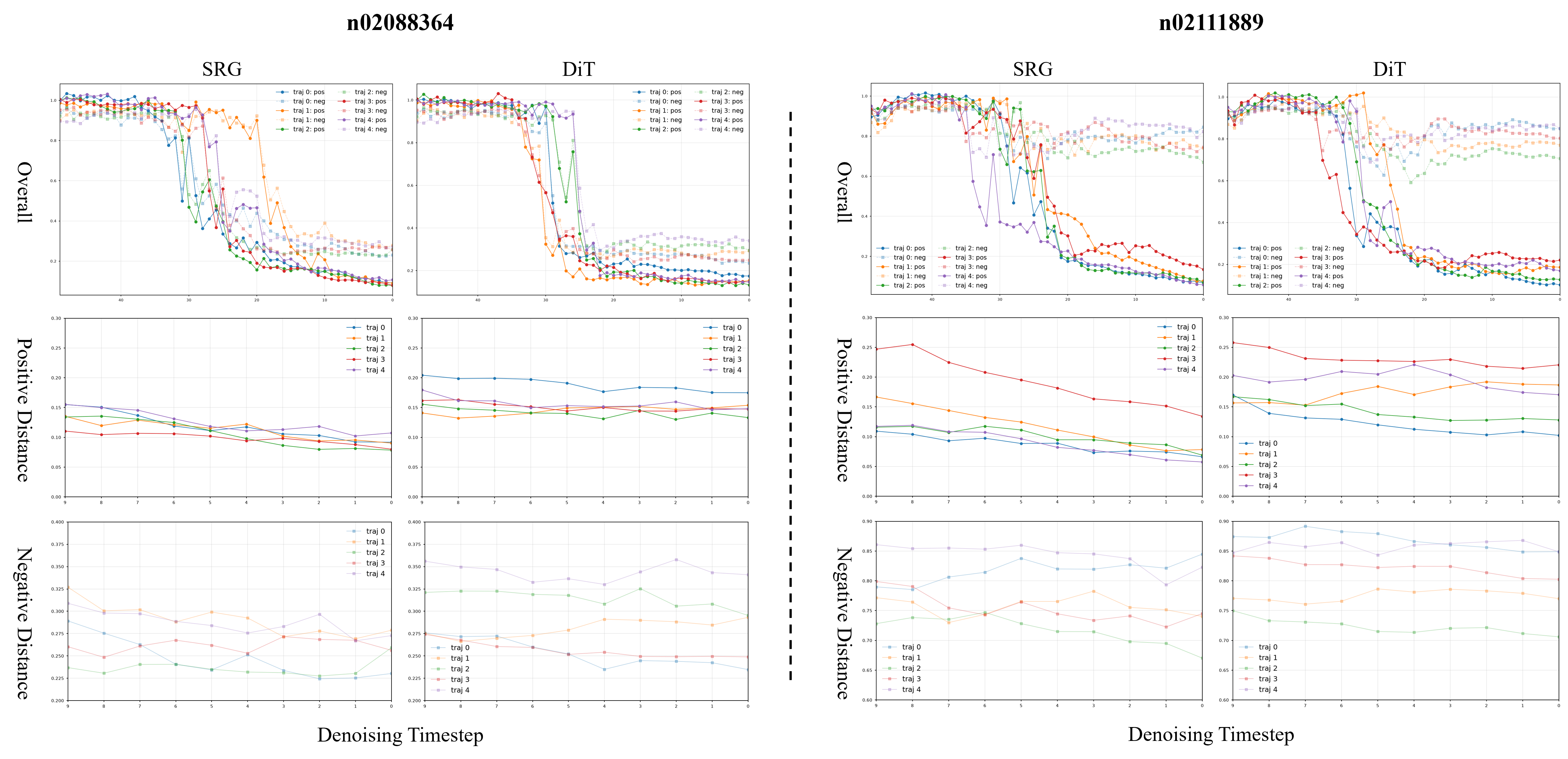}
    \caption{Evolution of prototype distances during denoising. The first row shows the complete trajectories. The second and third rows show the positive and negative distances, respectively, during the final $10$ denoising steps. The horizontal axis shows the denoising timestep, and the vertical axis shows cosine distance.}
    \label{fig:traj}
\end{figure*}
\subsection{Objective weights}
Since the overall strength of SSL guidance is already controlled by $\lambda_{\mathrm{SSL}}$, we only examine the relative weights of $\mathcal{L}_\mathrm{inter}$ and $\mathcal{L}_\mathrm{intra}$, using the following modified SSL-space guidance objective:
\begin{equation}
    \mathcal{L'}_{\mathrm{SSL}}^{i} = \mathcal{L}_{\mathrm{proto}}^{i} + w_{\mathrm{inter}} \mathcal{L}_{\mathrm{inter}}^{i} + w_\mathrm{intra} \mathcal{L}_{\mathrm{intra}}^{i}, 
\end{equation}
where $w_{\mathrm{inter}}$ and $w_{\mathrm{intra}}$ are relative weights. Because both the predicted representations and the prototypes are $\ell_2$-normalized, weights of $1.0$ place the three SSL-space objectives on comparable numerical scales. As shown in Table~\ref{sup:weight}, small weights yield behavior similar to that obtained using $\mathcal{L}_{\mathrm{proto}}$ alone, whereas excessively large weights cause the corresponding objective to dominate the guidance, interfere with prototype assignment, and degrade performance. This trend is consistent with the objective ablation: $\mathcal{L}_{\mathrm{proto}}$ provides the primary improvement, while $\mathcal{L}_{\mathrm{inter}}$ and $\mathcal{L}_{\mathrm{intra}}$ serve as auxiliary signals. We set $w_{\mathrm{inter}}=w_{\mathrm{intra}}=1.0$ to simplify the formulation while maintaining balanced performance.

\section{Out-of-Distribution Generalization}
\begin{table}[t]
\centering
\caption{Comparison of DiT, MGD3~\cite{chan-santiago2025mgd3}, and SRG on ImageNet-R. The best result in each IPC setting is highlighted in bold.}
\label{tab:r}
\begin{tabular}{cccc}
\toprule
Method & IPC=1 & IPC=3 & IPC=5 \\
\midrule
DiT  
& $58.6_{\pm 0.3}$ 
& $62.3_{\pm 0.3}$ 
& $62.6_{\pm 0.2}$ 
\\

MGD3 
& $56.0_{\pm 0.5}$ 
& $61.1_{\pm 0.2}$ 
& $63.5_{\pm 0.2}$ 
\\
SRG  
& \bm{$63.9_{\pm 0.3}$} 
& \bm{$64.4_{\pm 0.2}$} 
& \bm{$64.9_{\pm 0.3}$} 
\\
\bottomrule
\end{tabular}
\end{table}

To evaluate whether the distilled datasets preserve semantic information beyond the visual style of the source domain, we conduct a cross-style generalization experiment on ImageNet-R~\cite{Hendrycks2021imagenetR}. ImageNet-R contains 200 ImageNet classes rendered in diverse artistic and non-photorealistic styles, including cartoons, paintings, and sculptures, and therefore provides a challenging benchmark for robustness to style shifts. In this experiment, all distilled datasets are generated from the original ImageNet domain, while evaluation is performed on ImageNet-R.

\par

SRG consistently outperforms both DiT and MGD3 across all IPC settings. This result suggests that representation-guided sampling helps preserve semantic structure that remains useful under style shifts rather than merely improving visual realism in the source domain.

\begin{figure*}[t]
    \centering
    \includegraphics[width=\linewidth]{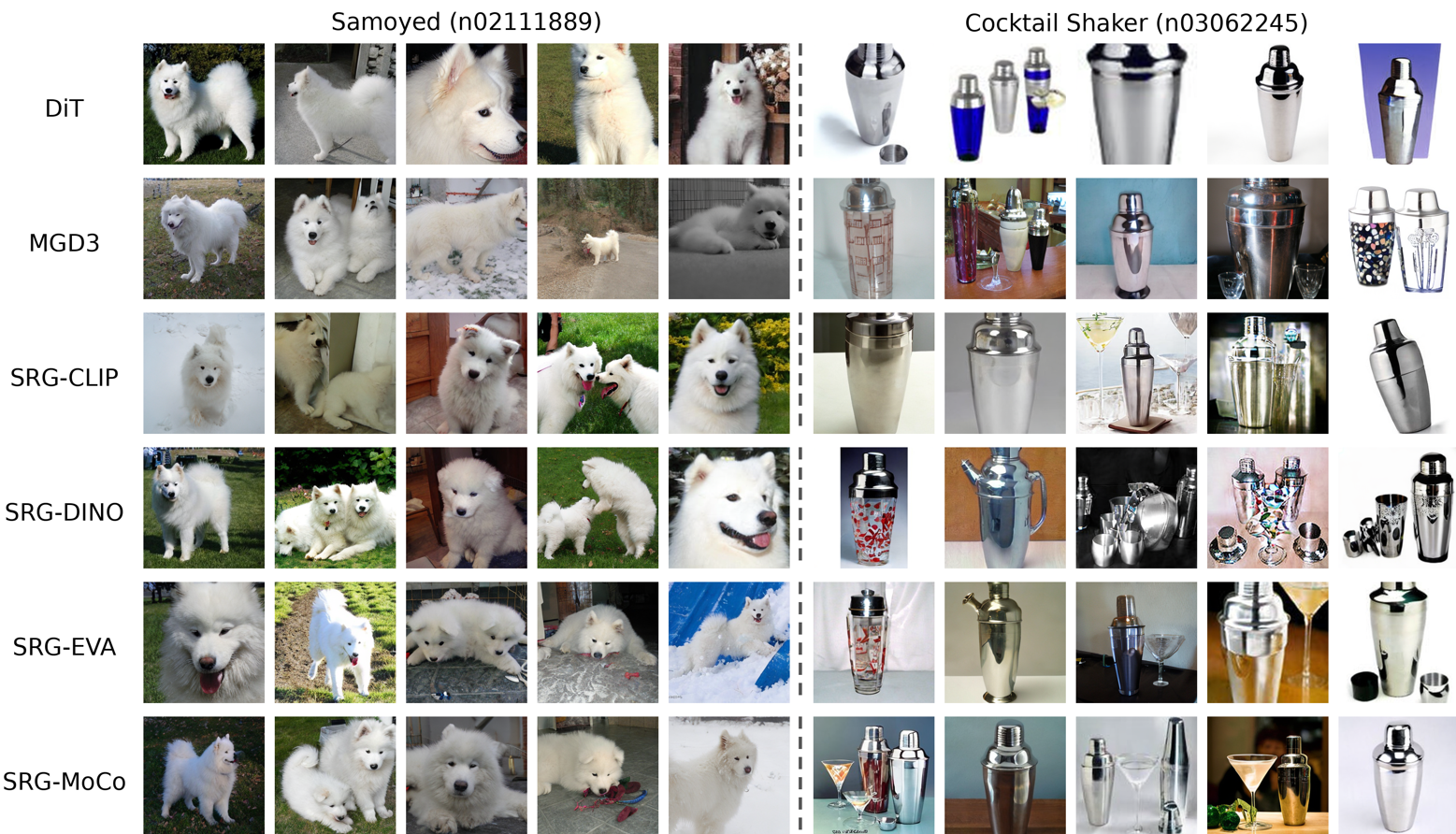}
    \caption{Samples generated by DiT, MGD3, and SRG with different SSL encoders for two representative classes at IPC $=5$. The generated samples exhibit different visual characteristics across methods and guidance encoders.}
    \label{sup:images}
\end{figure*}

\section{Robustness across Generation Seeds}

\begin{table}[t]
\centering
\caption{Performance of SRG on ImageNet-100 across different generation seeds. The Mean row reports the mean and standard deviation across the five seeds. The best result in each IPC setting is highlighted in bold.}
\label{tab:seed}
\begin{tabular}{cccc}
\toprule
Seed & IPC 1 & IPC 3 & IPC 5 \\
\midrule
0 & $89.0_{\pm 0.1}$ & $92.3_{\pm 0.2}$ & $\mathbf{93.2_{\pm 0.2}}$ \\
1 & $88.9_{\pm 0.2}$ & $92.3_{\pm 0.1}$ & $93.1_{\pm 0.2}$ \\
2 & $88.0_{\pm 0.2}$ & $92.1_{\pm 0.1}$ & $93.1_{\pm 0.1}$ \\
3 & $88.6_{\pm 0.2}$ & $\mathbf{92.6_{\pm 0.1}}$ & $92.7_{\pm 0.1}$ \\
4 & $\mathbf{89.1_{\pm 0.1}}$ & $92.3_{\pm 0.1}$ & $93.0_{\pm 0.1}$ \\
\midrule
Mean & $88.7_{\pm 0.4}$ & $92.3_{\pm 0.2}$ & $93.0_{\pm 0.2}$ \\
\bottomrule
\end{tabular}
\end{table}

We further evaluate the robustness of SRG across different generation seeds. As shown in Table~\ref{tab:seed}, performance exhibits moderate variation across seeds, especially under the most challenging IPC $=1$ setting. Nevertheless, SRG consistently maintains strong performance across all tested seeds. Compared with the generative baselines reported in the main results, SRG remains clearly superior even when seed-level variation is considered, indicating that its performance is not overly sensitive to the generation seed.

\section{Qualitative Analyses}

\subsection{Evolution of Prototype Distances}
We analyze the evolution of SSL-space representations throughout the denoising process for two representative ImageWoof classes: Beagle (n02088364) and Samoyed (n02111889). At each denoising step, we measure the cosine distance, defined as one minus cosine similarity, between the generated sample and its nearest prototype from the target class (positive distance), as well as its nearest prototype from a non-target class (negative distance). As shown in Fig.~\ref{fig:traj}, the first row presents the complete trajectories, while the second and third rows provide detailed views of the positive and negative distances, respectively, during the final $10$ steps. Compared with unguided DiT, SRG consistently moves the generated samples closer to target-class prototypes, as indicated by the lower positive distances, while retaining separation from the nearest non-target prototypes. These results confirm that SRG modifies prototype-based distances as intended and suggest that improved coverage of SSL prototypes can benefit downstream adaptation of the SSL model.

\begin{figure*}[t]
    \centering
    \includegraphics[width=\linewidth]{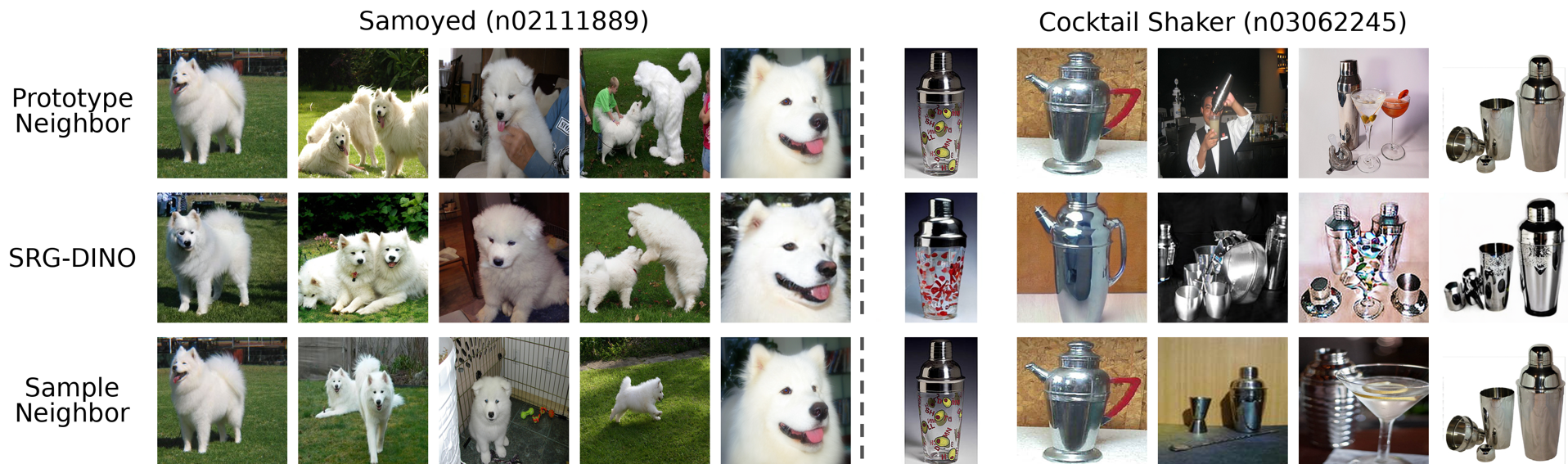}
    \caption{Comparison of prototype neighbors, SRG-generated samples, and sample neighbors in the DINOv2 representation space at IPC $=5$. Each column corresponds to one prototype.}
    \label{sup:neigh}
\end{figure*}
\begin{figure*}[t]
    \centering
    \includegraphics[width=\linewidth]{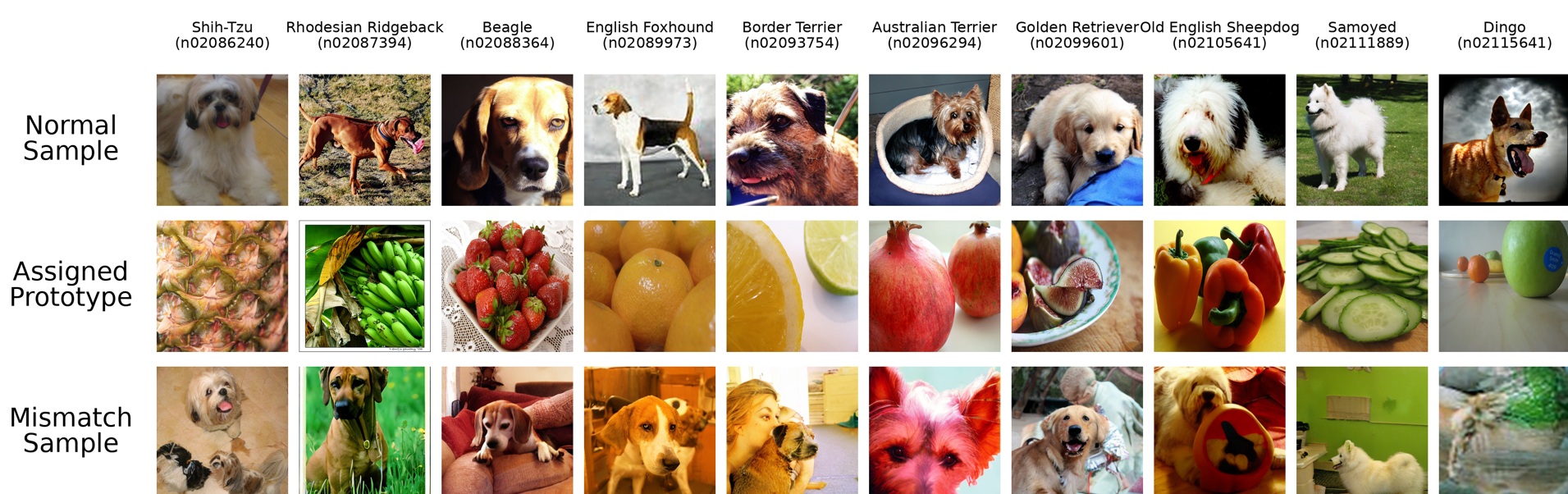}
    \caption{Comparison of samples generated with standard SRG guidance, mismatched prototype neighbors, and mismatched generated samples. The first samples generated using $\mathrm{IPC} = 5$ are illustrated.}
    \label{sup:mismatch}
\end{figure*}

\subsection{Distilled Samples}
Fig.~\ref{sup:images} presents distilled samples produced by DiT, MGD3, and SRG with different SSL encoders for Samoyed (n02111889) from ImageWoof and Cocktail Shaker (n03062245) from ImageNet-IDC at IPC $=5$. The unguided DiT samples exhibit relatively similar visual characteristics. In contrast, MGD3 and the SRG variants produce more diverse appearances, including variations in object shape, pose, texture, and background. The samples generated by SRG also differ across SSL encoders, suggesting that each encoder emphasizes different representational properties and consequently induces a different generation trajectory. Overall, the results indicate that the choice of guidance space affects not only downstream utility but also the visual modes captured by the distilled dataset.

\subsection{Comparison of Samples and Neighbors}
We compare each distilled sample with the real sample nearest to its assigned prototype (Prototype Neighbor) and the real sample nearest to the generated sample (Sample Neighbor). As shown in Fig.~\ref{sup:neigh}, SRG preserves the dominant visual attributes of the prototype neighbor in many cases without exactly reproducing it. This observation indicates that SRG balances SSL-space guidance with the visual prior of the pretrained diffusion model. An illustrative example appears in the fourth Samoyed column: the prototype neighbor contains a person wearing a dog-like costume, whereas SRG generates an actual dog with a similar low-level appearance but different semantic content. Overall, these results show that SRG integrates representation-space targets with the generative prior, producing samples that preserve structure around the assigned prototypes while remaining on a plausible visual manifold.
\begin{table}[t]
\centering
\caption{Comparison of downstream validation accuracy between standard SRG guidance and mismatched prototype guidance using $\mathrm{IPC = 5}$.}
\label{tab:wrong}
\begin{tabular}{ccc}
\toprule
Set & Woof & Fruits\\
\midrule
Woof  
& $92.0_{\pm 0.5}$ 
& $10.8_{\pm 2.5}$ 
\\

Fruits 
& $8.1_{\pm 1.8}$ 
& $89.1_{\pm 0.7}$ 
\\

Mismatch  
& $78.5_{\pm 2.1}$
& $72.2_{\pm 2.0}$
\\
\bottomrule
\end{tabular}
\end{table}

\subsection{Cross-Class Prototype Guidance}
To further examine the interaction between the generative prior and SSL-space guidance, we conduct a controlled intervention in which samples use class conditions from Woof but are guided by prototypes constructed from Fruits, and compare the generated samples (Mismatch Sample) with samples obtained using standard SRG guidance (Standard Sample), as well as the nearest neighbor to the assigned mismatch prototype (Assigned Prototype). As shown in Fig.~\ref{sup:mismatch}, the mismatch samples retain the overall shapes and object structures of dogs but exhibit atypical illumination and local details. We compare their downstream behavior with that of standard SRG samples in Table~\ref{tab:wrong}. When evaluated on Woof, the mismatched samples achieve lower accuracy than standard Woof SRG samples, yet attain substantial accuracy on Fruits. Such cross-dataset performance far exceeds that when standard samples are evaluated with another dataset. These results indicate SRG samples contain information associated with both the generator conditions that keep global visual identity, and the SSL space prototypes that remain discriminative semantic structure. 

\end{document}